\documentclass{article}

\usepackage{arxiv}
\usepackage[utf8]{inputenc} 
\usepackage[T1]{fontenc}    
\usepackage{hyperref}       
\usepackage{url}            
\usepackage{booktabs}       
\usepackage{amsfonts}       
\usepackage{nicefrac}       
\usepackage{microtype}      
\usepackage{lipsum}
\usepackage{graphicx}
\graphicspath{ {./images/} }
\usepackage{pifont}
\usepackage{amsmath}
\title{Intent Recognition in Conversational Recommender Systems}
	
\author{
 Sahar Moradizeyveh*$^{a}$\\
  $^{a}$Macquarie University, Sydney, Australia\\
  sahar.moradizeyveh@mq.edu.au\\
}

\begin{document}

\maketitle
 	
\begin{abstract}
Any organization needs to improve their products, services, and processes. In this context, engaging with customers and understanding their journey is essential. Organizations have leveraged various techniques and technologies to support customer engagement, from call centres to chatbots and virtual agents. Recently, these systems have used Machine Learning (ML) and Natural Language Processing (NLP) to analyze large volumes of customer feedback and engagement data. The goal is to understand customers in context and provide meaningful answers across various channels. Despite multiple advances in Conversational Artificial Intelligence (AI) and Recommender Systems (RS), it is still challenging to understand the intent behind customer questions during the customer journey.
To address this challenge, in this paper, we study and analyze the recent work in Conversational Recommender Systems (CRS) in general and, more specifically, in chatbot-based CRS. We introduce a pipeline to contextualize the input utterances in conversations. We then take the next step towards leveraging reverse feature engineering to link the contextualized input and learning model to support intent recognition. Since performance evaluation is achieved based on different ML models, we use transformer base models to evaluate the proposed approach using a labelled dialogue dataset (MSDialogue) of question-answering interactions between information seekers and answer providers. 
\end{abstract}

\keywords{Recommender Systems, Conversational Recommender Systems, Intent Recognition, Chatbot}

\section{Introduction}

This chapter begins with an overview of the approach and describes the problem statement. Our approach identifies the user's intent in chat-based conversational recommender systems from raw multi-turn conversational data. Then, we present our contributions and explain how the proposed new approach facilitates identifying user intent in conversation.

\subsection{Overview and Problem Statement}
Any organization needs to improve their products, services, and processes~\cite{benatallah2016process,beheshti2018iprocess}. In this context, engaging with customers and understanding their journey is essential. Organizations have leveraged various techniques and technologies to support customer engagement, from call centres to chatbots and virtual agents. Therefore, developing an intelligent conversational system using NLP, ML algorithms, and conventional interfaces attempts to assist customers and organizations through dialogue. 
Understanding customers' intent in a text-based interactive context and providing meaningful answers across various channels is a significant step to obtaining customer satisfaction. 

The fundamental idea of Recommender Systems (RS) is to offer users personalised promotions or suggestions on specific news or products~\cite{elahi2021recommender}. RS techniques typically focus on monitoring the users' behaviour, filtering data, and aiming to provide appropriate recommendations~\cite{beheshti2020personality2vec}. The conversational Recommender System (CRS) combines the sequential RS with the Dialogue System (DS) and proposes an integrated framework that represents two systems together. A human-machine conversational system emulates human verbal interactions to provide automated dialogue to help customers by answering questions, recommending items of interest, and performing complex tasks (e.g., in business decision-making).
Identifying the appropriate recommendation depending on the user query is an essential computational task of a dialogue system.
Chatbot-based CRSs face challenges in extracting intent and providing required actions for user requests. 

\begin{figure}[!ht]
    \centering
    \includegraphics[angle=0, scale=0.35]{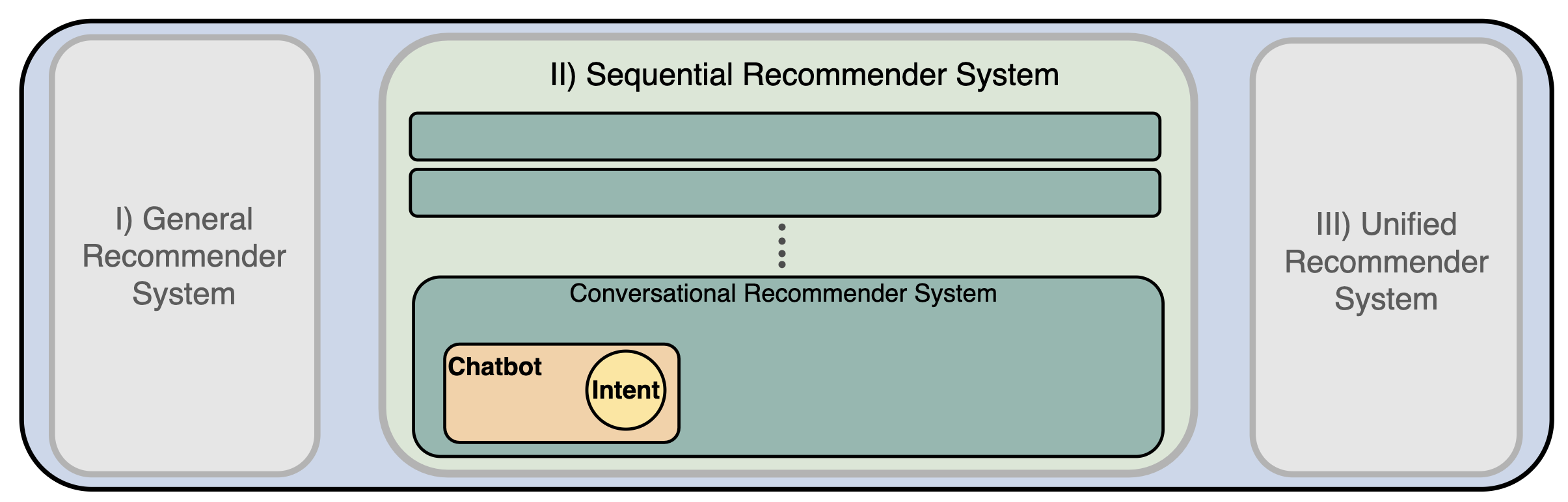}
    \caption{An overview of the work scope in Recommender Systems to identify intent in CRSs.}
    \label{fig:work-scope}
\end{figure}

Figure 1.1 shows the general overview of the research scope.
Hence, this paper first studies different RS types, including general, unified, and Sequential RS. Then we explore the CRS as a type of sequential RSs. Finally, we examine the text-based CRSs or chatbots to extract and recognize user intents.

First, we present a novel model to recognize the user intent in chatbot-based CRS to address these challenges. Next, we introduce a machine learning pipeline to contextualize the input utterances in conversation. Then, we utilize reverse feature engineering to collect technical data and redesign the strategy. Finally, we link the contextualized input and learning model to identify intent recognition. Performance evaluation is achieved based on different machine learning models and approaches. We used the advantages of transformer base models, which quickly became the state-of-the-art sequence-to-sequence NLP method, for this proposed approach. We evaluate our system with the MSDialogue dataset~\cite{chap1_1}, which is created for analyzing and understanding different aspects of the conversation. This data set is labelled by experts in the field and created by the Center for Intelligent Information Retrieval (CIIR) at the University of Massachusetts Amherst based on conversations with the technical Microsoft forums from 2005 until 2017.

\subsection{Contributions}
In this paper, we analyze various approaches in CRSs. We propose a novel model for the insight discovery process relying on transformer-based approaches. This model facilitates extracting intent in a multi-turn dialogue system to provide an accurate recommendation in the future. To achieve this goal, we present an interactive procedure.
The unique contributions of this paper are:

\begin{itemize}
    \item We introduce a pipeline to contextualize the input utterances in conversation. We customize existing data curation strategies to turn the user's conversational data into contextualized data.
    \item  We introduce a reverse feature engineering pipeline to collect unique data from the labelled databases through crowdsourcing and create a manual redesign strategy. 
    \item  We enable linking the contextualized input dialogue and transfer learning model to identify and recognize the user intents in multi-turn CRSs.
    \item We evaluate our approach with real-world datasets and highlight how the proposed model can help recognize user preferences in Web-based conversations.
\end{itemize}

\subsection{Summary and Outline}

    \begin{itemize}
        \item In Section 2, we  
        explain the current state-of-the-art approaches and summarise the techniques and methods in these areas.
        This chapter includes Recommender Systems, Conversational Recommender Systems, Chatbot-based Conversational Recommender Systems, and Intent Recognition in Conversational Recommender Systems.
        
        \item In Section 3, we present our methodology for using extracted intent from conversational systems to provide a more suitable recommendation. Meanwhile, this chapter presents precise information about the method pipeline, including data curation,  crowdsourcing for labelling data,  reverse feature engineering, contextualizing extracted intent, using the BERT classifier, linking the extracted result and providing feedback.
        
        \item In Section 4, we discuss the problem statement with the motivating scenario. Then, describe the experiment's dataset and setup and model classifier, demonstrate the experiment outcomes and conclude this section by presenting the evaluation results.
        
        \item In Section 5, we conclude the study, discuss the proposed method's next steps and present the project's future work.
    \end{itemize}

\section{Background and State-of-the-art}
\subsection{Recommender Systems}

Each second, we generate a considerable amount of data.
This data is collected in various forms; structured (e.g., a student record in an education system), semi-structured (e.g., a Tweet on Twitter), and unstructured (e.g., audio, video, image, and text files).
Moreover, a large amount of meta-data, i.e., information about data, could be collected by tracking the users' behaviour through rating, downloading, and tracing other activities such as following or sharing content over social media~\cite{beheshti2022social,beheshti2020istory}.

The basic idea of Recommender Systems (RS) is to offer users personalized promotions or suggestions on specific news or products~\cite{yakhchi2020enabling,yakhchi2018cnr,yakhchi2020towards}. RS techniques typically focus on monitoring the users' behaviour, filtering data, and aiming to provide appropriate recommendations.
The RS research domain is mainly related to Information Retrieval (IR)~\cite{chap2.1_1}, the theories about forecasting~\cite{chap2.1_2}, and cognitive science ~\cite{chap2.1_3}. Therefore, the recommender system is vital for content production and e-commerce media. Netflix\footnote{https://www.netflix.com/}, Amazon\footnote{https://www.amazon.com/}, Tinder\footnote{https://tinder.com/}, YouTube\footnote{https://www.youtube.com/}, and Facebook\footnote{https://www.facebook.com/} are famous representatives of companies among users in generating recommendations. 

Generally, RSs could be categorised into three primary categories~\cite{chap2.1_4}:
General Recommender systems (GRS), Sequential Recommender systems (SRS) and Unified Recommender Systems (URS).In the following sections, we present these three preliminary types of RS and review them in detail.
\subsubsection {General Recommender System}
    GRS aims to discover users' long-term preferences and could acquire this goal explicitly or implicitly by tracking historical users' activities. Hence, attempt to recommend the most relevant item and provide precise recommendations.
    Technically, GRS is divided into three main classes content-based, collaborative filtering, and Hybrid-based ~\cite{chap2.1_5}.
    In the following section, we will discuss these three classes of general recommender systems and review the current investigations in detail. 
    \bigskip
    \begin{itemize}
        \item \textbf{Content-Based Recommendations}
        
            Information retrieval(IR)~\cite{chap2.1_6} is the core idea of this type of RS. Users who had similar interests in past may have more similar preferences for future behaviours. Therefore, the Recommendation Engine(RE) provides a matching recommendation for users. Figure 2.1 demonstrates a basic form of a CF model, where the model recommends similar books relying on the user profile and previous interested items.
            
    \begin{figure}[!ht]
    \centering
    \includegraphics[angle=0, scale=0.35]{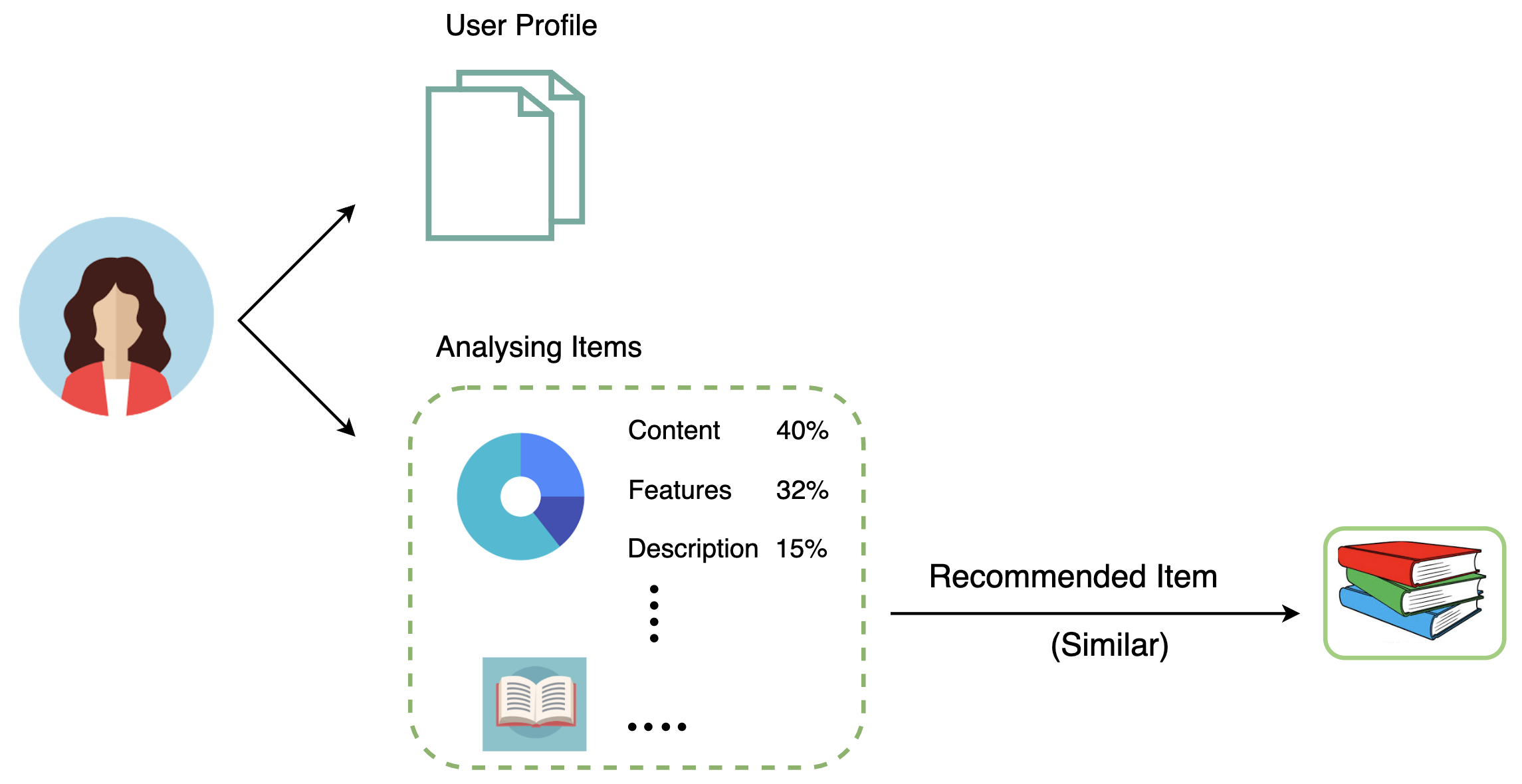}
    \caption{The extensible frame of content-based recommendations.}
    \label{fig:rsContent}
\end{figure}
            Content-based RSs generate a user profile by exploring available content from text, images, feedback, rates, likes/dislikes, and videos that a user focused on them previously. This measure identifies the best user profile and produces the most successful and accurate recommendation. 
            
            There is much research about content-based RSs in various domains. For instance, Re: Agent~\cite{chap2.1_7}is an intelligent email mechanism that can learn related actions (e.g., filtering, prioritizing, downloading, and forwarding emails). The Friend Of A Friend(FOAF)~\cite{chap2.1_8}  is a music recommender system that recommends relying on the user's interests. 
            Keyboard matching or Term Frequency /Inverse Document Frequency(TF-IDF) and Word2Vec(W2V)~\cite{chap2.1_9}are the typical approaches in this family of recommender systems. 
            In the past decade, deep learning (DL) has attracted much attention compared to conventional models, relying on its ability to deal with problems.
            For instance: Ask Me Any Rating (AMAR)~\cite{chap2.1_10} is a system that uses Long Short-Term Memory (LSTM) and provides recommendations based on the user's preferences.  
            
        \item \textbf{Collaborative Filtering Recommendations}
        
        For years, in many studies, Collaborative Filtering(CF) was interpreted as a replacement for RSs. A large portion of RSs restricted to CF. The main goal of the CF approach is to determine the most exciting items based on the similar reaction among users and point out more probability that users with the same preferences in the past have the same interests in the future~\cite{chap2.1_11}. Figure 2.2 shows a basic frame of collaborative-based recommendations.
        CF algorithms indicate better performance than the Content-based approaches, which depend on the user characteristics and item detail for making a prediction ~\cite{chap2.1_12}.
        
\begin{figure}[!ht]
    \centering
    \includegraphics[angle=0, scale=0.3]{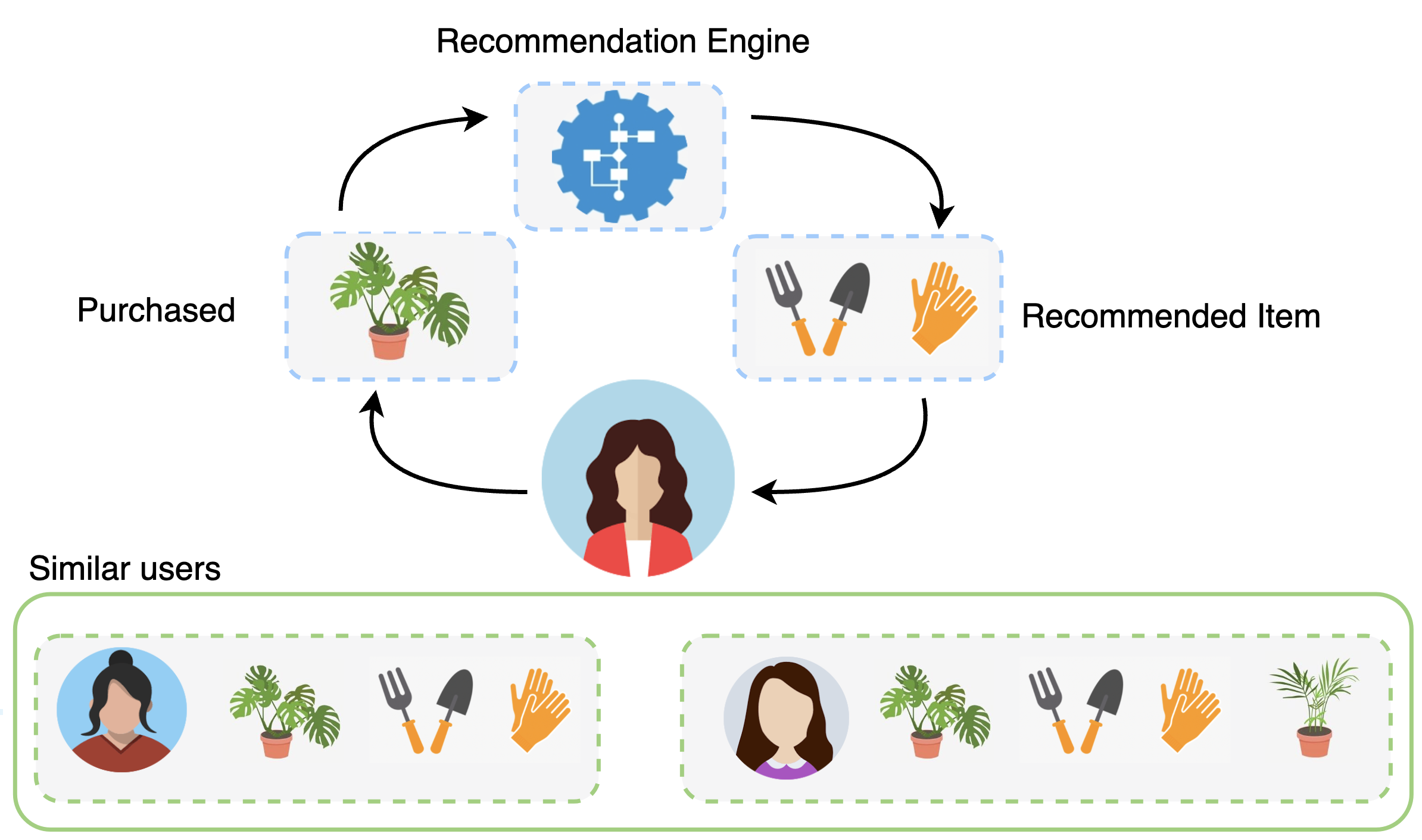}
    \caption{The extensible frame of collaborative-based recommendations.}
    \label{fig:rsCollaborative}
\end{figure}
     
        Goldberg et al. ~\cite{chap2.1_13} mentioned collaborative filtering by proposing the Tapestry in 1992. Tapestry is an electronic mail system that filters emails depending on the user's interests. This effective filtering system works similarly to the human process~\cite{chap2.1_14,beheshti2022ai}.
    
        A popular method in this class could mention in Matrix Factorization (MF). This Mathematical method is a simple embedding instance and tries to find the hidden relationship between users and items in various domains. Besides, MF is one of the best solutions for data sparsity problems. The MF approach works with implicit and explicit feedback types of data. The first one notes rating and review. The second one deals with browsing via clicking and the history of purchases. Over the years, different CF classifications have been presented in the literature, which can be categorized into one of the main classes: memory-based and model-based approaches~\cite{chap2.1_15}.  
        \begin{enumerate}
             \item\textit{Memory-based }or Neighbourhood-based approach, compare a peer user-item interaction and, based on historical behaviour, develops recommendations and works over the whole database for predictions. Can mention to GroupLens ~\cite{chap2.1_16} and Ringo ~\cite{chap2.1_17} as a memory-based collaborative- filtering RSs. GroupLens is a news filtering system that helps people to find enjoyable news. This architecture predicts news scores via Better Bit Bureaus and tries to collect and extract the ratings. Ringo is a dynamic music network system that tries to construct personalized recommendations.
             
            \item\textit{Model-based} by performing algorithms on various massive databases, try to respond to the user instantly. This class of approaches deals with the relations between various items to build the model based on the connection and then provides the top suggestions for the target user. With the rapid progress in machine learning, different methods such as clustering models ~\cite{chap2.1_18}, rule-based, and dimensionality reduction techniques ~\cite{chap2.1_19} were proposed.
        \end{enumerate}

        \item \textbf{Hybrid-based Recommendations}
        
            This class of RSs efforts to improve the accuracy of RSs by using collaborative filtering and Content-Based recommendation. There are three main forms for the combination of CF and Content-Based recommendation methods~\cite{chap2.1_20}:
        \begin{itemize}
            \item Calculating the recommendation via both methods in separate ways. The absolute result relies on the combining rates from both CF and content-based recommendation techniques, or it will generate from which that has more accurate results~\cite{chap2.1_21}.
            \item  Integrating collaborative filtering or content-based method characteristics into the other one.
            Fab~\cite{chap2.1_22} is an editorial assistance system. The first version of Fab was proposed in 1994. Fab is one of the first hybrid systems to use both approaches' benefits.  
     	    \item Add individual RS to the next one as a single strategy and use the benefits of both models.
            As we discussed previously, the Collaborative Filtering Recommender Systems(CFRSs) perform better than the content-based, considering a couple of reasons~\cite{chap2.1_23}: i) Without information about the user profiles, they can provide recommendations. ii) They do not require complex Natural Language Processing (NLP) methods to analyse item descriptions.
    \end{itemize}
\end{itemize}

\subsubsection {Sequential Recommender System}

    Indicating long-term user preferences is the principal task of general recommenders. While these classic approaches can model users' interests, they suffer from learning short-term interests.
    Therefore, GRSs can not recognize the changing patterns of individual interests over time.
    SRS aims to model sequential behaviours by awareness of the short-term preferences of the user. They only consider the recent actions of the users and record them in their accounts. Hence, the user can act anonymously without logging into the platform, and there is no coercion to recognize the users by the system.
    
    SRS emerged as a new approach in recent years. SRS operate particular computational tasks by assuming different problem characterisations.
    This class of the RS caused research on various platforms such as music, movie, and browsing recommendation.
    According to the current studies on SRSs, we can categorise them based on different characteristics such as interactions, models, and inputs~\cite{chap2.1_24}.

\bigskip

\begin{itemize}
    \item \textbf{Characteristics}
In many scenarios, users have different tastes and characteristics (e.g., in e-commerce, companies can not find a specific pattern for their shopping behaviours). Hence, the industries may face various challenges, which below we mention to them:
    \begin{itemize}
        \item \textbf{Handling Noisy Sessions in User-Item Interaction:} 
            Traditionally, we can not discover a specific pattern for the user's decision procedure. (e.g., the user may change their mind at the last second, which may cause a different decision). It is a critical issue that has not been studied enough.
    
            However, by using the attention mechanism, some investigations show the contribution and relations of items to have a better idea about the user interest and subsequency to increase the recommendation's performance. The Attention-based transaction embedding model (ATEM) ~\cite{chap2.1_25} was proposed in 2018. This model measures the weight of each observed entity in a transaction without considering the order.
            \medskip
        \item \textbf{Distinguish and Follow a Session Purpose:}
            In some cases, due to the user's curiosity, they may involve the irrelevant items in sequential actions and construct a model based on the incorrect dependencies, leading to unsuitable recommendations. Accordingly, distinguishing the user's intent is vital for a sequential RS. For example, the user starts browsing about cake equipment to purchase and then tries to calculate and find a replacement item instead of the sugar. This curiosity may direct the user in search of different objects and causes irrelevant user-item interactions.
            Neural Attentive Recommendation Machine (NARM)~\cite{chap2.1_26} is a framework that tries to model the user's sequential behaviours to understand the purpose of the interaction session.
            \medskip
        \item \textbf{Handling Order Dependency in a Session:} 
            In reality, the products purchased in a particular transaction are unrelated or in order(e.g., first selecting the flour or baking powder does not matter, or even inserting an unrelated item like pasta into the list).
            According to the present studies~\cite{chap2.1_27}, there is little awareness of this uncomplicated observation. The current methods include Recurrent Neural Networks(RNN), factorisation machines, and Spatial Predictive Models(SPM), which do not satisfactorily extract the dependence relationships and sequential patterns. On the other hand, ruled-based approaches are a good alternative for modelling adjustable interaction sessions. Ruled-based method predicts the following recommendation via learning the most frequent pattern in the interaction sections. For instance, the E-Learning framework ~\cite{chap2.1_26} was proposed in 2010, which authorises a flexible combination of components in the recommendation.
            \medskip
        \item \textbf{Handling the Long User-Item Interaction Session}
        Dealing with long-term sequential interaction between a user and an item and modelling their dependencies are other challenges for modelling sequential recommender systems.
        To address long-term dependency, different architectures such as the Markov chain, various types of RNNs (e.g., Long Short Term Memory(LSTM)~\cite{chap2.1_28}, Gated Recurrent Unit (GRU) ~\cite{chap2.1_29}, and attention mechanisms attempt to provide better performance.
    \end{itemize}
    \bigskip
    
    \item \textbf{Input Data}
    Sequential recommender systems (SSRs) are classified into four general groups according to input data:
    \begin{itemize}
        \item \textbf{Only User-item Interaction Session:} 
        This is the elementary input data type and includes a couple of recorded actions in a sequential recommender.
        \medskip
        \item \textbf{Single/Multiple Behaviour User Data:}
         In SSRs, a range of actions to determine the user interest level are considered.
        For instance, we can mention to review, click on advertisements, place items in the purchasing basket or wishing list, and search for specific keywords. 
        Assuming a couple of actions simultaneously for the input data is challenging for a sequential recommender system that may cause a performance reduction in the recommendation~\cite{chap2.1_30}. 
        \medskip
        \item \textbf{Auxiliary information:} 
        Side information is the complementary details that drive the more accurate recommendations and learning of the context of the transactions required to understand the user's intents and interests~\cite{beheshti2019towards}. For this purpose, the Neural Network-based transaction Embedding Model (NTEM) was proposed in 2007 ~\cite{chap2.1_31}. The model attempts to learn the items-embedding and the related transaction features in parallel. Eventually, illustrate the result of evaluation metrics for the efficiency of the recommendation.
        \medskip
        \item \textbf{Repeat Purchasing Behaviour:} 
        Purchasing similar items is typical behaviour in real-world scenarios. For instance, brand loyalty or identical life habits lead to repetitive behaviours. RepeatNet  ~\cite{chap2.1_32} relies on the proposed attention mechanisms to choose the best option at the right time.
    \end{itemize}
    \item \textbf{Model Structures}
    \bigskip
    Similar to different aspects of Artificial intelligence, new challenges in sequential recommendation caused the investigation of more complex and integrated algorithms. These algorithms could be classified into eight different models:

    \begin{itemize}
    \item \textbf{Sequential Pattern Mining:}
    SPM is a data mining method~\cite{chap2.1_33} and made noteworthy progress among various domains (e.g., e-commerce~\cite{chap2.1_34, chap2.1_35})
    SPM detect the repetitive patterns in user-item interaction and tries to discover the next-items recommendation. (e.g., if most users prefer to buy a phone case after purchasing a new mobile phone, that is a good idea to suggest it).
    \item \textbf{Pattern/Rule-Based Approaches:}
    The idea of this approach is to determine the frequent behavioural pattern in user-item interaction without considering the item's order based on the three notations: (i)session matching, (ii)item suggestions, and (iii)systematic pattern mining.
    For instance, the model proposed by Mobasher et al. ~\cite{chap2.1_36} is a personalised method that relies on capturing data via clickstream in web server logs to provide a sufficient recommender framework.
    \item \textbf{Markov Chains Models:}
    CM is a straightforward method among the behaviour analysis approaches, based on the hypothesis that the existing state depends only on the previous state or the few last interactions~\cite{chap2.1_37}. This approach is delivered in two subsets, the basic Markov chain-based and latent Markov embedding-based. The first one mention the condition that the probability of each state is related only to the previous state and only model the uncomplicated dependencies between two states. The second one tries to calculate the Euclidean distance between states to extract more complex relationships.
    \item \textbf{Factorised Machine-Based Models:}
    With the emergence of the Recommender Systems, Matrix factorisation was a basic approach for many models in this field and overall maps the user-item interaction. Factorized Personalized Markov chains (FPMC)~\cite{chap2.1_38} is a sample of this category. FPMC combines the factorisation and Markov chains to learn the user's historical behaviour and the sequential features among baskets.
    
    \item \textbf{Neural Network-based Models:} 
    Neural Network-based models have had remarkable achievements in natural language processing in recent years. Convolutional Neural Network-based, Graph Neural Network-based, and Attention Mechanism-based Models are the pioneering models in this area.
    
    In 2018, the Convolutional Sequence Embedding Recommendation Model (Caser) \cite{chap2.1_39,yakhchi2022convolutional} was presented as the first \textit{Convolutional Neural Network(CNN)} model. Caser predicts the top-N ranked items to the user. First, it saves all previous user interactions- items in a matrix, then assumes horizontal and vertical filters, and finally provides these matrices into a CNN network as an image.
    
     \textit{Recurrent Neural Network(RNN)} model has a significant role in sequential recommender systems and natural language processing and leads to the proposition of novel models in this field. Following this, an approach like ~\cite{chap2.1_40}, by characterising the user profile(short and long-term preferences), tries to indicate how RNNs can achieve better predictions. 
    
    Recently, increasing attention on the \textit{Graph Convolutional Network (GNN)} ~\cite{chap2.1_41}, as an approach to learning graph node embedding, especially in natural language processing, caused remarkable research in this field. A structure Aware Convolutional Network (SACN) is a pioneering example of GNN approaches proposed in 2019. SACN is an end-to-end convolutional neural network, which includes an encoder, decoder, node attributes and edge relation types. It contains adjustable weights based on the acquired information from neighbours that lead to having accurate embeddings nodes.
    
    In 2019, the Google brain team caused a revolution in the literature by introducing the \textit{Attention Mechanism}. Since then, most proposed models have been according to the attention mechanisms. The most influential capability of the Attention mechanism in comparison to other methods, especially the RNNs family, is that it can work on selective parts of the data. Attention-based models accept the input data without hierarchy and deliver output data simultaneously. The attention mechanism illustrates noteworthy results in dealing with noisy data.
    Vanilla ~\cite{chap2.1_42}, as a preliminary example of this type of neural network, is used widely in sequential recommender systems.
 \end{itemize}
\end{itemize}
\subsubsection {Unified Recommender Systems}          
    As mentioned in previous sections, the CRSs address the long-term users’ preferences and the SRSs target short-term interests. The emergence of the third model created a new framework by combining both GRSs and SRSs approaches. Hybrid-based RSs improve the accuracy of the recommender systems by using the advantages of both previous methods~\cite {chap2.1_43}.
    
    Overall, the hypothesis behind each recommender system is to provide (i) valid suggestions, (ii)satisfy users’ preferences,(iii) a quick response, (iv) increase the business profit and (v) facilitate users with their decision-making process~\cite {chap2.1_44}. Therefore, providing accurate and swift responses is the essential task of the RSs and will increase industrial profits. Accordingly, companies try to consider diverse user interests, which include short-term and long-term preferences. For instance, A student always has a scientific book in her preference basket as a long-term preference, but on holidays, she/her prefers to read a romance book, which could be considered a short-term preference. 
    
    Although analysing the short and long-term preferences in separate ways is more straightforward, recent research indicates that the combination of both types of users’ preferences in a unique framework gains superior performance. For example, Factorizing Personalized Markov Chains (FPMC)~\cite {chap2.1_45}is one of the basic Unified frameworks that use the Markov chain approach to achieve sequential user behaviour. This method could not deal with a few issues, such as data sparsity and long-tailed distribution. However, it has acquired excellent outcomes compared to MF and MC.
\subsection{Conversational Recommender System}

The traditional recommender systems or static recommendation models ~\cite{chap2_95} with a one-shot/turn interaction strategy might lead to various limitations in distinguishing user preferences from current preferences. For instance, recommending a new laptop, based on the previous observation of the user's behaviours or dependent on the context, can cause a challenging situation to indicate the current user conditions, or even users need coaching to make the final decision. The emergence of Conversational recommender systems(CRS) as a type of Conversational Artificial Intelligence is considered an evolution in this field and helps to address many of these issues.

Early research on the CRSs back to the late 1990s when AutoTutor was proposed by Graesser et al.~\cite{chap2_4}. Indeed, AutoTutor as a Virtual Instructor (VI) attempts to simulate dialogue with students. AutoTutor assists students in different fields, such as cognitive science, psychology, and teaching and learning technology~\cite{barukh2021cognitive,schiliro2020cognitive,jalayer2020attention,schiliro2020novel,schiliro2022deepcog}.

Conversational Recommender System (CRS), as a sequential recommender system, is the combination of the Recommender Systems (RS) and Dialogue System (DS) and proposes an integrated framework that represents two systems together. Accordingly, This class of recommendation assume different approaches and covers various styles of system-user interactions. Through the years, interactions of user-machine reliance on context from the utterances have become more complicated. Humans-System interactions are present by different approaches, such as voice/speech-based (e.g., Alexa, Google Home, Siri, and Bixby) and text-based conversational systems, also known as chatbots, and used by different names such as chatter-bots, virtual agents, or dialogue systems and Conversational Assistants.
Over time, Conversational agents have a significant acceptance by users. Hence, increasing human expectations in this field have complicated human-machine interactions. Accordingly, extracting and predicting the user's intents is more challenging and causes significant interest in this class of Recommender systems.
 
\bigskip

Section 2.2.1 presents the characteristics of CRS. 
Next, section 2.2.2 conclude with various Interaction Approaches in CRSs. In Section 2.2.3, we explain the computational tasks in CRSs.
Section 2.2.4 discussed the various architecture of the CRSs.
Section 2.2.5 represents Knowledge and Data in CRS, and Section 2.2.6 describes interaction Modes in CRS.

\bigskip

    \subsubsection{Characterizations of CRS}

    Conversational systems have two main features, task-oriented and multi-turn. CRSs extract the user intent and preferences to provide the most relevant recommendation in natural language processing (NLP).

\begin{itemize}

    \item \textbf{Conversational Turn}:

    To maintain the user-machine interaction, the agent has to explicitly or implicitly extract data from dialogues. Meanwhile, keep the user passionate and provide precise responses. The conversational Recommender System deals with user needs and preferences for a sustainable period via voice, typed text, forms, or gestures. In text-based CRS, the machine can answer user inquiries in a simple one-turn question answering(QA session) or other questions that need details and more conversation turns. 
    
    Figure 2.3 shows Single-turn and Multi-turn conversations. To maintain the interaction, the agent has to explicitly or implicitly extract data from dialogues and keeps the user enthusiastic by providing precise responses.  
    \bigskip
    
  \begin{figure}[!ht]
    \centering
    \includegraphics[angle=0, scale=0.33]{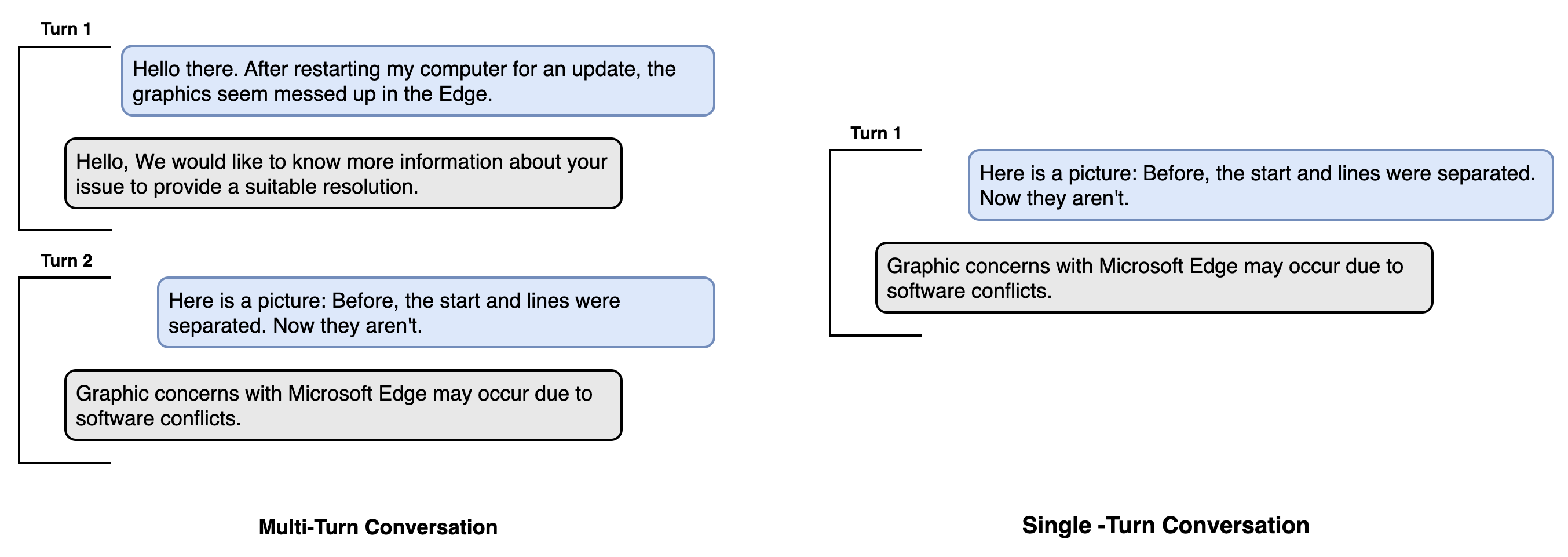}
    \caption{Multi-turn and Single-turn Conversation.}
    \label{fig:Conversation}
\end{figure}

    \item \textbf{Conversational Task}:
    \medskip
    
    Regarding the recommendation domain, conversational agents could be classified into task-oriented (goal-oriented) and non-task-oriented conversations. Although a simple conversation can be legislated even through handwritten notes, designing sophisticated conversational systems require deep neural networks to extract the user intent. Many dialogue-based approaches are concentrated based on goal-oriented dialogue systems. Selecting the appropriate procedure to understand users’ requests to perform a particular task and interact with humans via natural language is the critical stage of CRSs.
    
    Despite the task-oriented conversational system that provides a fascinating conversation, they are far from understanding and responding to the out-of-scope conversation. 
    This type of platform usually focuses on one duty at a time. This kind of conversational system assists the user with specific tasks in different domains such as online shopping~\cite{chap2_96}, healthcare ~\cite{chap2_97}, and education ~\cite{chap2_98}.

    Non-task-oriented conversational systems talk with users continuously. The non-task-oriented system is not restricted to a specific domain or purpose. Since the conversation is a two-way interaction, tedious responses quickly terminate the conversation. Hence one of the vital components of this system is user engagement. There is no doubt that a system with the ability to handle non-task-oriented with the accuracy of task-oriented systems is a desirable scenario. However, there is not much research on this situation yet.
\end{itemize}
\bigskip

    \subsubsection{Interactions in CRS}
\begin{itemize}
    \item \textbf{Interaction Modes}
  
Nowadays, interacting with a conversational agent is intertwined with people's daily lives. People use this technology regardless of age, from kindergarten to age care.
Interaction between users and agents could be accomplished via text or voice. Recently, \textbf{\emph{Voice-based}} conversational systems are becoming increasingly popular. Intelligent voice-controlled assistants such as Alexa, Google Home, Siri, and Bixby have made a comfortable life for many. Voiced-based conversational recommenders mention collecting data by asking questions through a conversation ~\cite{chap2_5}. According to ~\cite{chap2_6}, by the end of 2024, the number of voice assistants will be higher than the world's population. 

\textbf{\emph{Text-based}} conversational systems, also known as chatbots, and introduced with different names such as chatter-bots, virtual agents, or dialogue systems ~\cite{chap2_9} emerged in the late 1960s by Weizenbaum.~\cite{chap2_7}. This type of messaging platform provides natural contact with a system agent. Chatbots are used in conversational recommender systems to assist humans in decision-making in various aspects of their lives.

\item\textbf{Interaction Approaches}
    
    Early conversational assistance tries to analyse the syntaxes via natural language. Rule-based or decision-tree agents are based on the hierarchy of rules to provide the solution for different types of problems. By mapping the utterances with the assumed assumptions, they predict the oncoming question and the related responses. Overall, they expect the next stage based on the trained scenario. Ruled base systems are more affordable, more rapid to train, have backward compatibility with previous generations, and are not limited to text interactions.

    In the late 1970s, Elaine Rich ~\cite{chap2_1} proposed a mechanism named stereotypes, based on two supposed different scenarios that tried to interact with users via questions in natural language. In the earliest 1980s, RABBIT, a brilliant database assistant based on the critiquing methodology, was welcomed~\cite{chap2_2}. This framework is based on the psychological hypothesis and tries to provide the queries. 
    
    In contrast to the Rule-based system, which has a slightly flexible conversational flow, there are learning-based models~\cite{tabebordbar2020feature,tabebordbar2020adaptive,tabebordbar2018adaptive}. The learning-Based or Deep Learning-Based dialogue systems van analysed according to the models and system types.       
    
    \item\textbf{Application Environment}     
    
       CRS has two main application environment categories: the stand-alone and embedded environment. In addition, regarding the environment, the supported devices (e.g., in the voice-based recommendation, specifically designed hardware, help in voice-based interactions.) significantly impact selecting design approaches.

    \begin{enumerate}

    \item \emph{Stand-alone environment:} The recommendation is the main concentration of the Application. For example, ~\cite{chap2_106} is an intelligent web-based framework based on the domain-independent repository that presents the tourist consultative. The Wasabi Personal Shopper (WPS)~\cite{chap2_107} is another example of the stand-alone FindMe system with a self-dependent dataset developed for searching electronic product catalogues based on text or queries online.  
    \bigskip
    
    \item\emph{ Embedded environment:} In this type, the CRS is not assumed as an independent system and is embedded as part of the framework in the format of a chatbot~\cite{chap2_108}, voice-based assistants ~\cite{chap2_109} or even as part of 2D or 3D user experience ~\cite{chap2_110}. Hence, the recommendation is one of the capabilities of such systems.
    \end{enumerate}
  \end{itemize}  

\subsubsection{Data and knowledge in CRS}

Following the rapid increase in data generation, concerns such as information overload and redundant information issues have had significant importance in various fields in recent years~\cite{beheshti2019datasynapse,yang2021design}. The main challenge here is the cognition overload of information that needs to be processed for the recipient of the information \cite{zakershahrak2018interactive, zakershahrak2020online, zakershahrak2021order}.
Correspondingly, the Conversational Recommender Systems are not an exception from this critical point. Meanwhile, understanding the natural language and providing a meaningful response requires a knowledge base. Hence, this part examines different data and knowledge as the significant aspects of CRSs.

\begin{itemize}
    \item \textbf{Data}
    Structured layouts (e.g., forms in web-based applications), Natural Language based (e.g., in written or spoken assistance), and Application-Specific are three main approaches that support the inputs and outputs of data~\cite{coredb}.
    In the first approach, the machine follows a task-specific path and tries to interact with humans by choosing from pre-prepared forms. The natural language-based approaches, on the other hand, relies on natural language interactions and frequently provide more flexible solutions. Task-oriented conversation systems ~\cite{chap2_102}, Deep learning-based systems ~\cite{chap2_100}, or  Audio to the text approaches ~\cite{chap2_101}-are examples of these types of forms. The list-based form relies on a natural Language dialogue system~\cite{chap2_103} in the simple example of the Hybrid approach, which combines the two mentioned models.
    However, few applications accept unique categories of input and output data (such as geographic maps~\cite{chap2_105}, three-dimensional space ~\cite{chap2_104}).

 \item \textbf{Knowledge}
    In the 1980s, Knowledge-graph (KGs) presented as the understanding of knowledge from different resources \cite{chap2_17,beheshti2020intelligent,beheshti2016scalable}. Since this definition had a restriction in numerable relations and likewise suffered from the lack of common characteristics, the emergence of a new illumination by Google in 2012 improved the comprehensive depth of knowledge graphs\cite{chap2_161}. The recent description emphasizes semantical annotation by interpreting the generated relation between objects and actions. The acquired knowledge is one of the significant factors in generating the recommendation. This knowledge can obtain explicitly or implicitly from different resources, such as supported user intents by CRSs, user modelling, dialogue states, and item-related information (background knowledge)\cite{chap2_18}.

    \textbf{i) Background knowledge} or prior knowledge mentioned to understanding the origin of the related subject and in CRSs helps to learn the language of the conversation. Background knowledge has an influential role in continuing the flow of the conversation. This knowledge could be extract base on four main categories:
    
 \begin{enumerate}
	\item{ Logged Interaction Histories}: Analyse the utterances in conversation. Understanding the user's requirements, even in basic steps, can provide a better understanding of the user preferences or feed back~\cite{chap2_111}.
    \item{ Dialogue Corpora Created to Build CRS }
    Dialogues based on natural languages usually use data from different annotated conversations provided by crowdsources (e.g., Amazon Mechanical Turk). For instance, in ~\cite{chap2_111} where Sun et al. use the collected conversation from live dialogues and generate the recommendation.
	\item{ Item-related Information}: Include general explanations, any provided tags and ratings, comments or other metadata (e.g., the item attributes or descriptions) that could store in the related database.
	\item { Lexicons and World Knowledge}
    using different knowledge bases is a strategic way to recognize entities in NLP-based approaches(e.g., Wikipedia, WordNet, Wikiquote).
\end{enumerate}

\textbf {ii) User intent:} The dialogue system based on information filtering about specific purposes provides the best-related recommendation. Hence, extracting the user intents, especially in NLP-based systems, is another crucial element for suggesting the acceptable option.
There are several approaches in the literature to identify the user's intent and choose appropriate responses to them. Built-in intents are a straightforward way for this purpose. In such systems, intents rely on the manual engineer, and background knowledge could be pre-defined. However, supported intents according to the domain requirements could be different.

The intent is classified into two main groups:  domain-independent ~\cite{chap2_113} and system-supported~\cite{chap2_114}. Domain-independent user intents include initiating the conversation (greeting), questions (different types of questions, for example, clarifying or general questions, etc. ), answers, utterances with no useful information(junk utters), obtaining more information (details and explanations), feedback (positive, and negative feedback).
\bigskip

\textbf{iii) User modelling: }Generally, user modelling (UM) is the process of using the collected data from a user to improve the system's knowledge ~\cite{chap2_115}. In CRSs, user modelling refers to different modalities and interactive strategies for understanding user Preferences.
There are a couple of approaches to representing the extracted user preferences which can classify based on different conditions:
Short or long-term~\cite{chap2_119} preferences, engineering strategy~\cite{chap2_117}, system manner~\cite{chap2_118}, and situation with cold-start ~\cite{chap2_116}.
\bigskip

\textbf{iv) Dialogue status:} Defining the dialogue state to choose the following appropriate action helps to have a stable chat in multi-turn conversations. Therefore, in many CRSs, Dialogue Management (DM) is critical in dialogue flow and conversation progress. DM strategies show the road map for the following action/response in the conversation. Generally, Dialogue Management can be classified into two main approaches: Rule-based and statistical methods (based on machine learning methods). DMs use different types of data, for instance, the history of the previous dialogue, domain knowledge, and spoken language understanding~\cite{chap2_20}.
DM strategies show the road map for the following action/response in the conversation. Finally, attention to the interaction and confirmation strategy are two main factors that we have to consider in building DMs. 
\end{itemize}

\subsubsection{Computational Tasks in CRS}  
Tasks in CRSs placed in two main categories: principal and supporting jobs. Principal Tasks are the spinal cord of a CRS and include requesting, recommending, explaining, and responding.
\bigskip

\begin{itemize}
    \item \textbf{Principal Tasks}
    
    \emph{Request:} During the last decades, various procedures to distinguish the preference information regarding items or attributes were proposed. In addition, this methods mostly try to specify the oncoming question to increase the efficiency of the conversation. For instance, in some approaches, they try to create a pre-defined structure to understand the sequence of possible questions that is famous as  Entropy-based methods~\cite{chap2_122, chap2_123}. For instance, eliciting food preferences in \cite{chap2_10}, or  fashion sense recommendation in \cite{chap2_11}.
        
    \emph{Recommendation:} A recommendation offers the best possible and relevant options base on the domain. The three primary recommendation categories are explained more in section 2.1.
        
    \emph{Explain:} Providing relevant information in order to explain the recommendation is another task of CRS, and various papers attempt to analyze it\cite{chap2_3}. Efficiency, effectiveness, persuasiveness, transparency, and satisfaction are five different factors considered for appropriate advice\cite{chap2_13, ghodratnama2021summary2vec}. 
    
    \emph{Respond:} This category practically refers to users' questions and provides reasonable responses. Recognition of the pre-defined intents(e.g., Understanding intents in online Shopping~\cite{chap2_14}) and generating the responses by different machine learning algorithms( e.g., for personalizing in dialogue by training the neural networks ~\cite{chap2_15}) are two main approaches to provide the answers.
        
    \item \textbf{Supporting Tasks}
    
    CRS has several supporting tasks relying on the system's functionality that can mention in Natural Language Understanding (NLU), Sentiment Analysis, and Specific Recommendation Functionality(SRF).
    
    \emph{Natural Language Understanding}
    The primary purpose of NLU is to enable the machine to interact with a human without supervision. Two essential notions of NLU are intent recognition (IR)~\cite{chap2_124}, and entity recognition (ER)~\cite{chap2_125}. NLU tries to analyze and choose the appropriate meaning of the data via algorithms, and IR focuses on extracting specific sentiments from the input text. Meanwhile, ER  is a process of determining the entities and extracting information about them \cite{ghodratnama2021intelligent}.
    
    \emph{Sentiment Analysis} is the process of understanding the approximate sentiment of the user's feeling about the object based on the rating, like and dislikes.
    For instance, ~\cite{chap2_126} is a specific framework based on movies, music, and book domains as a CRS, with three interaction methods natural language, buttons, and a combination.
    
    \emph{Specific Recommendation Functionality} is another supporting task in CRSs, helping to provide proficient suggestions based on the choosing recommendation technique \cite{ghodratnama2020extractive}.
    For instance, ~\cite{chap2_127} introduce a generative probabilistic model to produce context-aware recommendations for the famous and long tail queries.
  
\end{itemize}
\subsubsection{Architecture of a CRS}
    Typically in CRSs, different architectures and related components propose based on the system functionality. Computational Elements and knowledge Elements are two main categories of architecture. Figure 2.4 illustrate the typical architecture of the CRS. based on the ~\cite{chap2_128} with more details.
    
    \begin{figure}[!ht]
    \centering
    \includegraphics[angle=0, scale=0.5]{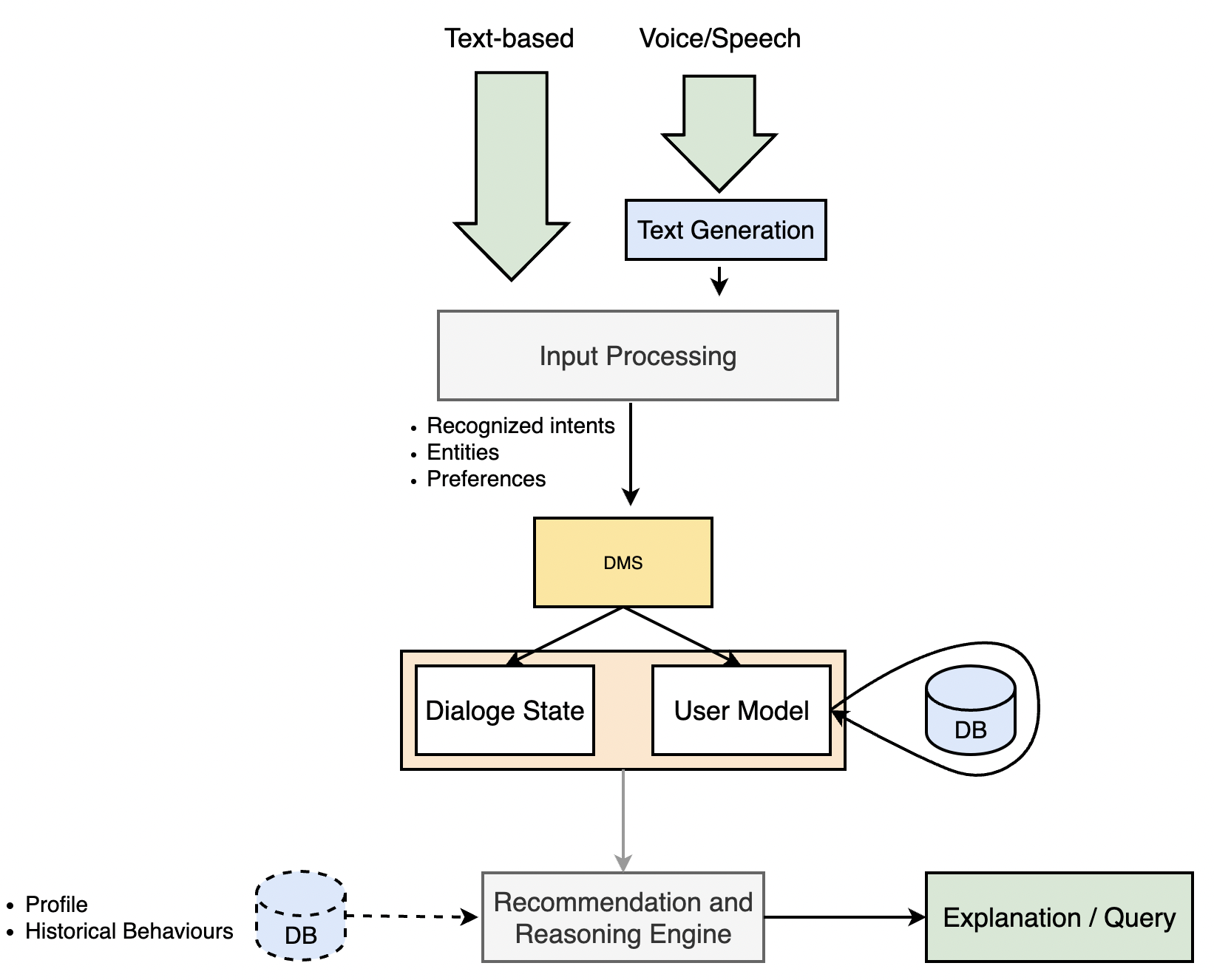}
    \caption{Typical architecture of the CRS. based on the ~\cite{chap2_128} by inserting more details. }
    \label{fig:TypicalArchitecture }
\end{figure}

    \begin{itemize}
        \item \textbf{Computational Elements}
        The \emph{Input/Output Modules} are the first and last part of the procedure in an architecture.
        The input processing based on the Natural language input types (e.g., speech, text) extracts the user's intent and item attributes and tries to identify the intent and entity recognition~\cite{chap2_129}.
            
        The \emph{Dialogue Management System(DMS)} or \emph{status tracker} is a fundamental part of the technical architecture in CRS and manages the processing flow from the input to the output results~\cite{chap2_132}.
            
        The \emph{User Modeling System(UMS)} is part of DMS that could be an independent component, especially when the system is looking for long-term preferences. UMS created based on the calculated interest weights regarding any features of an item. UMS use this inference knowledge to provide the related recommendations~\cite{chap2_131}.
            
        \emph{Recommendation and Reasoning Engine (RRE)}, based on the current user's preferences, has a role in retrieving and generating descriptions of the available recommendations in the system.
        \item \textbf{Knowledge Elements}
        
        \emph{Item Database} represent the suggestions with details about the attributes.
            
        \emph{Domain and Background Knowledge}, the dialogue knowledge can be explicitly Embedded in the forms of the pre-defined conversations and the possible transfer data among the dialogue states or can automatically learns from the dialogue or other domain sources~\cite{chap2_130}.
    \end{itemize}

\subsection{Chatbot-based Conversational Recommender System}
    A chatbot is an online framework focused on interacting with humans based on natural language and can prepare appropriate responses relying on knowledge from different domains. Chatbots transform human-machine interactional patterns. During the last decades, chatbots have obtained great attention as an outstanding aspect of Artificial Intelligence(AI), and various research has been performed in this field. This intelligent conversation with a human tries to simulate human behaviour in the conversation. This conversation could perform in text or voice/speech.
    The progress in chatbot technology over the years changed companies' policies to attract and communicate with their customers.
    Hence, this dialogue agent is an inseparable part of the growth in different domains such as e-commerce\cite{chap2_31}, education\cite{chap2_32}, finance\cite{chap2_33}, health\cite{chap2_34}, entertainment \cite{chap2_35}, news \cite{chap2_36}, and productivity \cite{chap2_37}.
    Figure 2.5 shows a standard simplified scheme of a chatbot-based recommender system.
   
\begin{figure}[!ht]
    \centering
    \includegraphics[angle=0, scale=0.32]{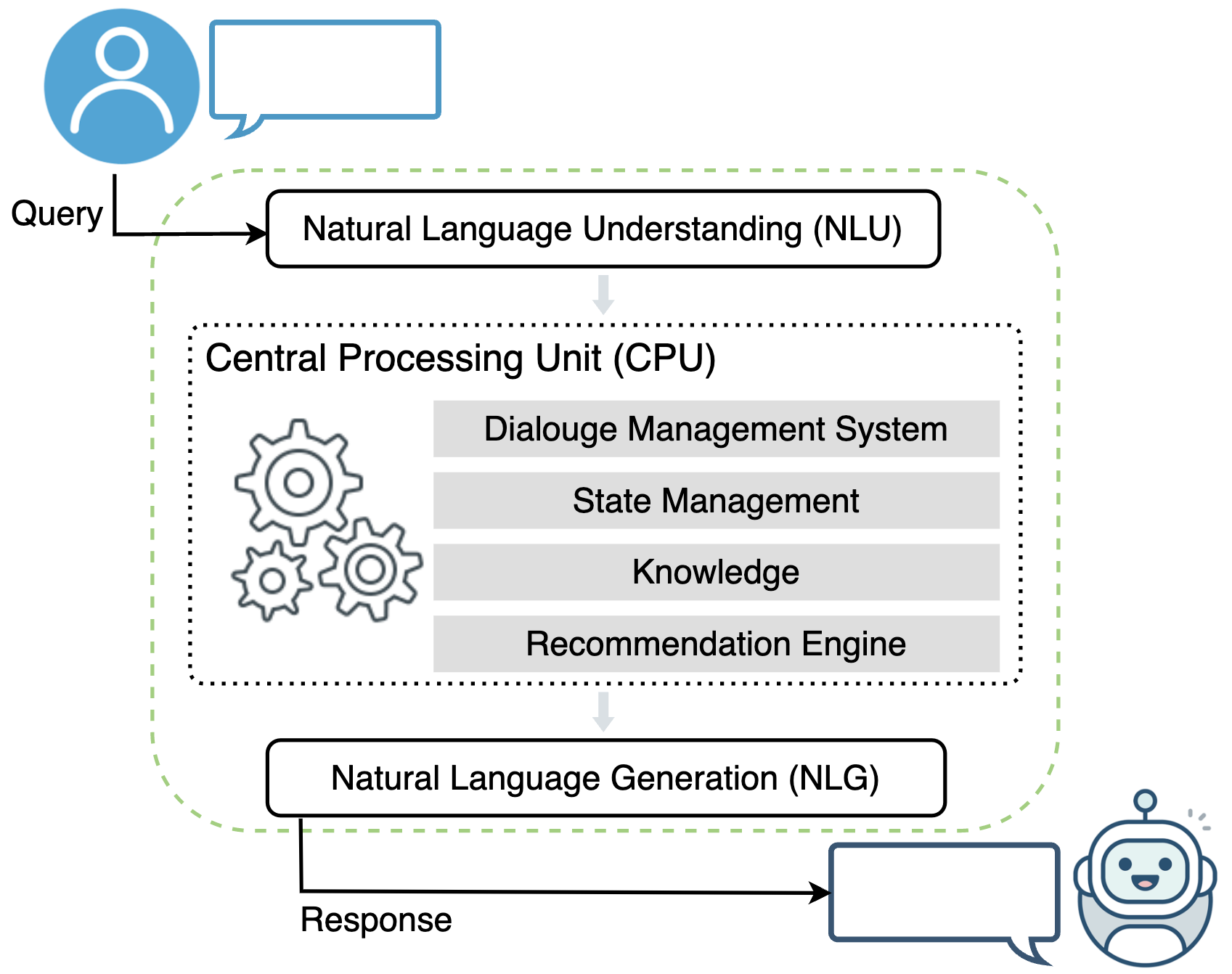}
    \caption{General view of a chatbot-based CRS}
    \label{fig:chatbot}
\end{figure}
\bigskip

    Conversation processing occurs in three main steps: Natural Language Understanding(NLU), Central Processing Unit(CPU), and Natural Language Generation (NLG).
    The processing unit manages the flow of the process, while NLU and NLG communication interface between humans and machines.


\subsubsection{Background and History}
    According to the literature, Alan Turing ~\cite{chap2_29}, asking "Can machines think?" mentioned the idea of a chatbot in the 1950s. Early chatbots like Eliza~\cite{chap2_7} focus on interaction with humans via natural language. Eliza tries to talk with the user based on the keyboard matching and identifying the context of the conversation.
   
   In 1972~\cite{chap2_30}, Artificial Linguistic Internet Computer Entity(ALICE) was proposed by Kenneth Colby. This new emergence was significantly more complex than the earliest ones.ALICE uses pattern matching and stores extracted data in the knowledge base. Chatbot technology has improved dramatically with advances in natural language processing and machine learning until today by famous voice-based and text-based chatbots such as Alexa, Siri, and Cortana~\cite{chap2_30}.

\subsubsection{Classifications of chatbot} 
Regarding the design technology and the field of chatbots, four main criteria can classify within the field ~\cite{chap2_8}:
\begin{enumerate}
    \item \textbf{ Interaction Modes}
    Interacts with users and provides thriller conversations over their requirements can perform via text or voice.
    \begin{itemize}
        \item\textbf{Voice-based Chatbot }
        These days, chatbot assistance (like Siri or Alexa) is an inseparable part of people's lives (e.g., helping them with scheduled behaviour tracking ). Accordingly, with growth in popularity, expectations from voice-based chatbots are much more elevated. 
        \item\textbf{text-based chatbot} This type of agent provides an enriched form of communication in various domains with humans through natural written language. A chatbot tries to understand user needs and simplify user interactions. Generally, this type of communication is integrated with various text-based communication media (e.g., messaging apps and social media). Hence, the agent must have the ability to obtain accurate intent based on the conversation and provide solutions that require customers to type or tap their requests and commands.
    \end{itemize}
    \item \textbf{Domain}
    \begin{itemize}
        \item \textbf {Open-domain Chatbot:} is an agent with the ability to continuously talk about different topics in natural language. 
        Producing this agent is quite challenging in various aspects 
        , and to explore this type of conversational chatbot, broad research has been done on it, such as Meena ~\cite{chap2_38}, MILABOT ~\cite{chap2_39}, Mitsuku~\cite{chap2_172}, and Cleverbot~\cite{chap2_171}, though a significant unsolved concern remains.
        All of these chatbots, with diverse frameworks, try to demonstrate human-like attributes.
        \item \textbf{Domain-specific Chatbot:} or closed-domain chatbots focus on providing related responses that rely on specific intent or keywords to perform tasks based on specific domain knowledge. Most service-based chatbots are a subset of this group ~\cite{chap2_40}.
    \end{itemize}
    \item \textbf{Design approaches}
    \begin{itemize}
        \item \textbf{Rule-based approach:} This type of chatbot works according to simple or complicated pre-defined rules. Therefore, this agent cannot discover the answer to any question outside its boundary~\cite{chap2_136}. Rule-based chatbots are valuable for environments with specific search engines and reduce the additional burden on the workforce ~\cite{chap2_137}.
        \item \textbf{Retrieval-based Approach:} A chatbot based on the input utterances retrieval provides the most informative and clear answer from the pre-defined response repository.
        Preliminary works on retrieval-based chatbots concentrate on single-turn with short-text conversation ~\cite{chap2_134}. This type of chatbot provides answers based on a real-world conversation database. Besides this straightforward approach, other studies focus on the multi-turn conversation~\cite{chap2_135}. In comparison, this approach faces challenges like providing a response compliant with the context and declaring the connection among utterances.
        \item \textbf{Generative-based Approach:} This method is well-known as Self-Learning as well. The principle idea of this method is to develop a new conversation based on training a large dataset. This approach overcomes the restriction of previous models relying on the combination of different learning methods such as supervised, unsupervised and reinforcement learning~\cite{chap2_143}.
    \end{itemize}
\item \textbf{ Interaction Approaches}
    \begin{itemize}
    \item \textbf{Task-Oriented Chatbot:} Nowadays, Task-Oriented applications are part of our daily life. Siri and Alexa are task-oriented interactions and help users in specific scenarios in a specific domain. For example, in banking sector~\cite{chap2_144}, education ~\cite{chap2_145}, customer services ~\cite{chap2_146}, E-commerce and etc. These chatbots have access to specific domain knowledge.
    \item \textbf{Non-Task-Oriented Chatbot:} Refers to a system that  
    does not have any exact purpose or specific task. Non-task-oriented chatbots are designed to simulate unstructured interaction same as human-human conversation. Generative-Based and Retrieval-Based are two main classes of this architecture~\cite{chap2_148}. Since this type of conversational agent needs to respond based on the discussion flow, their design algorithm is more complex. Hence, they should be able to retrieve information from online sources.
    \end{itemize}
\end{enumerate}
\subsubsection{Chatbot Design Techniques}
    Rule-based, Retrieval-based and Generative-based Approaches operate based on different techniques, and in this part, we mention the most important ones.~\cite{chap2_8, chap2_24}:
    \begin{itemize}
        \item \textbf{Parsing:} This technique performs the following steps: (i) Extract meaningful information from input data. (ii) Convert that received text into more simple words. (iii) Stored, (iv) Manipulated them based on the requirements. ELIZA was the first chatbot that benefited from this method.
        \item \textbf{Pattern Matching:}  Pattern matching is a fundamental technique in many programming languages, similarly is a commonly used strategy in question-answering chatbots. Although early chatbots like Eliza used a simple form of this module, new modern chatbots employ the advantages of this model in more complex ways. Flexibility to generate conversations and scaling problems that cause extracting of repetitive responses are advantages and disadvantages of this technique, respectively.
        \item \textbf{AIML:} Artificial Intelligence Markup Language(AIML) is directed from Extensible Markup Language (XML). The main goal of AIML is to develop a flexible conversational flow in chatbots.
        Generally, AIML-based chatbots are in the rule-based chatbots category. Generally, AIML-based chatbots are in the rule-based chatbots category. A chatbot can have groups of AIML with different functionalities. An AIML-based chatbot can perform various texts with identical meanings as the input. Correspondingly, a chatbot can reply with a default statement if the input pattern is not satisfied~\cite{chap2_150}.
        \item \textbf{Chatscript: } Was a scripting language as an open-source, rule-based engine in 2010. Chat script language won the Loebner prize for a couple of years. This engine has a more efficient structure compared to AIML for pattern matching. The chat script cornerstone relies on the C++ programming language. 
        \item \textbf{Artificial Neural Networks Models:} Artificial neural networks (ANN) are a sequence of algorithms that attempts to imitate the human brain~\cite{chap2_152}. ANN has enabled the development of intelligent chatbots. Structural techniques proposed in ANN-based strategy are the main differences compared to rule-based chatbots. ANN-based conversational models bots can develop relying on both retrieval and generative approaches to respond ~\cite{chap2_151}. This type of chatbot uses different algorithms for modelling conversational behaviours that can mention in the recurrent neural network (RNN) and sequence to sequence and Long Short-Term Memory networks (LSTMs) ~\cite{chap2_153}.
    \end{itemize}

\subsection{Intent Recognition and NLP }

In the past decade, having dialogue assistance to provide customized recommendations has become more common among companies. These dialogue systems are created to implement specific objectives for decisions making based on information filtering.

As chatbots grow, they face challenges such as extracting intent and providing required actions for user requests. Identifying the appropriate action depending on the user query is an essential computational task of a dialogue system. Therefore, having a smooth conversation and constructing a trust-able interaction with the agent for a long time plays a significant role in building a conversational agent.

 \subsubsection{Domain-Independent Intent Classification in CRS}
   Traditionally, in domain-specific CRSs, the set of user intents is pre-defined and depending on the system's needs and conditions, they support a different set of user intents.
    Relevant research on user intent in the conversational agent is insufficient. However, the number of common user intents is available in most CRSs. The primary common user intents in the literature are identified as follows:
\begin{itemize}
    \item \textbf{Initiate, quit, restart the conversation:}
    In NLP-based CRS, matching the system strategy, the agent or the user can start the conversation and begin with a greeting.
    In some scenarios, the user may begin the conversation session from scratch and reset the previous dialogue. Ultimately, at the end of the conversation, regardless of user satisfaction, the conversation could be done or redirect the user to some specific forms( e.g., a user feedback statement)~\cite{chap2.4_3, chap2.4_4}.
    \item \textbf{Chit-chat:}
    Research shows that, in many conversational systems, chitchat is considered a vital element of the machine user experience. Hence, supporting chat and assuming it is the common goal of the user is an inevitable part of CRS~\cite{chap2.4_5}.
    \item \textbf{Request for Explanation and details:}
    Asking for recommendations may occur by beginning the conversation or even after receiving details or explanations. 
    In many situations, the user prefers to know more about the available items or even wants to compare similar items.
    Therefore, learning about details is an integral part of a system.
    On the other hand, the request for more explanation could occur from the agent and the user to provide more information.
    \item \textbf{Preferences and Recommendations:}
   As discussed earlier, user preferences take many forms and understanding user preferences is a critical factor for support recommendations.
    
    \item \textbf{Feedback:}
    User feedback is an outstanding source of information for RS to learn more about the system and provide better suggestions based on the customer experience and satisfaction. Feedback in different scenarios is collected in different ways and has significant value for the improvements of the RSs. User satisfaction may be positive or negative. Feedback is gained during the dialogue or through a questionnaire at the end of the conversation.

\end{itemize}


\subsubsection{Open Intent Classification in CRS}

Understanding user intents from written language is vital in providing automated responses. Study in this area decomposes into two main categories: the specified predefined intent investigation (closed-world classification) and open intent (open-world)discovery. 
Most current research relies on predefined intent and attempts to place the utterances in a restricted list of purposes. Managing open intent classification is a significant challenge, and extracting correct intent is meaningful in conversational systems.

Existing models for recognising the open intent in a preliminary phase try to find the decision boundaries(DB). However, there is no specific model to determine the DB for open intents, and the current studies seek to design new classifiers.
Scheirer et al. (2013) ~\cite{chap2.4_7} proposed a hybrid model that
estimates the probability of the different object classes to determine the open intent classification.
Hendricks et al. (2017)~\cite{chap2.4_8} introduce a new baseline based on NLP and computer vision to estimate the probability of the distribution and detect the out-of-distribution examples.
Zhang et al. (2021)~\cite{chap2.4_6} present a technique to discover the adaptive decision boundary (ADB) to identify the open intent classification.

\subsubsection{Predicting User Intent Scope}
An input query in conversational system is classified into \emph{(i) out-of-scope} and \emph{(ii) in-scope queries}
An out-of-scope query in a domain-specific refers to an unsupported question in the system. Since users are not aware of the restriction and capabilities of the chatbot, this type of query is not unavoidable. Hence, determining outliers is one of the challenges in such a system.

In 2019, Larson et al. ~\cite{chap2.4_1} identified three basic strategies for predicting out-of-range queries by training a large amount of data. 
First, they determine the query types.
Second, if it was in scope, they placed it into one of the pre-determined intents. 
In the last stage, they proposed baseline procedures to predict the out-of-scope prediction. This baseline procedure includes: training a further number of specified out-of-scope intents on training data and setting a threshold for the classifier’s probability. Eventually, results show models like BERT~\cite{chap2.4_2} obtain better performance on in-scope classification.
\subsubsection{Natural Language Processing}

    Natural language processing (NLP) is a data-driven field for understanding, manipulating and producing human language by extracting information from textual data through statistical and probabilistic estimation. NLP shows significant potential in decision-making in various industries.
    
    Advanced conversational systems operate using NLP and Machine Learning (ML) algorithms. In conversational-based systems (e.g., Siri), the system interprets the user's speech via different NLP algorithms and then provides the appropriate responses based on ML algorithms.
    The role of NLP is evident in processing and analyzing text and converting text or speech into valuable insights. Hence, considering the undeniable importance of NLP in dialogue-based systems and understanding the user's intent in the conversation, we will explore this field in detail.
    
    Traditionally, NLP classifies into \emph{core} and \emph{application} areas~\cite{chap2.4_9}. Based on this classification, many diverse fields (e.g., conversational systems, machine translation and question-answering systems) use NLP depending on their requirements.
    
    \bigskip
    
    \textbf{Core Area of NLP}
    \begin{itemize}
    \item \textbf{Language Modelling:} Refers to predicting the probability of words in the sentences, depending on the syntactic and semantic relationships between them.
    Language modelling is a crucial element of any NLP application and is vital in understanding the user's intent in many fields, such as machine translation, text summarization, and speech and text recognition.
    
    \item \textbf{Syntactic Processing:} Well-known as parsing is another area of research in NLP. Natural language is easy to understand for humans, while interpreting and understanding the meaning of words, phrases and sentences are complex tasks for machines. Syntax analysis refers to the relation between words and phrases in a sentence. Constituency and dependency parsing are two distinct forms of that.\emph{Constituency Parsing} mention the process of composition of the sentences base on the grammatical category in a hierarchical manner. \emph {Dependency Parsing}  investigates the connections and relationships between pairs of separate words. Dependency parsing is the most commonly used method in deep learning models. 
    Graph-based and transition-based dependencies are two subsets of dependency parsing. The first approach constructs a few parse trees and searches to find the relationship between phrases and words. 
    The second approach contains a buffer and a stack. The buffer has all the words of the sentence and the root labels of the words. Meanwhile, in the stack, the connections (known as arcs) are made between the top two items until the buffer becomes empty~\cite{chap2.4_10}.
    \item \textbf{Semantic Processing}
    Semantic analysis tries to extract the exact meaning of the sentence. Therefore, in the first step, analyze the purpose of each word and then elicit the intention of words in sentences by combining the words in the next step.
    
    \item \textbf{{Morphology}}
    A morpheme is the least meaningful linguistic thing in a language. 
    Inflection, derivation, composition and integration of the morphologies construct words. Having a suitable morphological analyzer is necessary for most NLP tasks. In NLP, morphological parsing is the procedure of determining the structure of the word, which contains the stem and various types of affixes. The morphologically aware language modelling ~\cite{chap2.4_11} and unsegmented morphological languages model ~\cite{chap2.4_12} are instances of this type of language model.
    \end{itemize}
    
    \textbf{Applications of NLP}
\bigskip 

    Fundamental of the applications area are associated with topics distinguishing, translation of the text, summarization, automated response, and differentiated documents via classification and clustering. Hence, numerous companies have launched their conversational agents. This trend has led many researchers in the information retrieval (IR) and natural language processing (NLP) community to pay more attention to conversational search. 

\begin{itemize}
    \item \textbf{Information Retrieval}
    Information retrieval (IR) refers to the chain of tasks to find the most relevant information in a suitable format from text-based data.
    The IR system (like Google Search) helps the user to find the most relevant response by exploring through a collection of natural language records without trying to discover or generate the answers.
    \emph{indexing} and \emph{matching} are two leading methods in IR systems. Indexing includes tokenization, removing frequent words and stemming. Matching is the procedure of discovering the similarity of distinct represented text.
    
    \item \textbf{Text Classification} Refer to a method to allocate the pre-defined labels to raw data. Text Classification has a significant role in many NLP tasks, such as sentiment analysis and topic labelling. Likewise, numerous models, methods and evaluation metrics have been presented for text classification in the literature. 
    Overall, related features could be extracted through traditional methods or automatically by deep learning methods~\cite{chap2.4_16}.
    Traditional methods such as ~\cite{chap2.4_17}, Naïve Bayes (NB) that discuss a technique for automatic indexing of documents based on their subject content; or support vector machines (SVM) ~\cite{chap2.4_18} to investigate the particular properties to classify the text data.
    \item \textbf{Text Generation}
    try to generate different texts, such as poetry or jokes relying on the input or background knowledge.
    Generating the poetry is the most challenging part of these applications ~\cite{chap2.4_19}.
    
    \item \textbf{Information Extraction} 
    An early IE was created in the 1970s ~\cite{chap2.4_15} and is simultaneous with the appearance of NLP. IE elicit and process structured information from unstructured and semi-structured context.    
    
    \item \textbf{Summarization}
    
    It aims to extract the automatic procedure to find the best representation of raw data and skim quick and straightforward through the context. In the summarized document, three main facts are considered:
    (i) A summary is made of one or more documents. (ii)The summarized document contains valuable facts about the context. (iii) The summary should be less than 50 per cent of the original document.
    
    Generally, there are two primary categories for summarization, extractive and abstractive. \emph{Extractive summarization} concentrates on extracting information from the documents and tries to simplify and change the order of the text.\emph{Abstractive summaries} focus on developing the abstract based on the new words. ~\cite{chap2.4_13}. Furthermore, text summarization can classify depending on various criteria such as input sources, output properties, and purpose of the Summarisation~\cite{chap2.4_14}
    
    \item \textbf{Question Answering}
    Refers to systems that try to locate a proper automatic response for requests by human's natural language.
    The early QA system was \emph{Baseball}~\cite{chap2.4_22}, as a computer program that provides a related answer to the questions in the standard English language regarding the stored data. This program used the same techniques as ELIZA, the early chatbot. The architecture of the question-answering systems has the significant progress over time, and in 2001, determining the kind of question and the answer was provided by the question classifier module~\cite{chap2.4_23}. Generally, QA systems are classified depending on the domain, data sources, and examination of the questions and related answers~\cite{chap2.4_24}.

    \item \textbf{Machine Translation} converts a document to another language automatically. MT provides fluent translation as output using mathematical and algorithmic techniques. In the 1950s, as the demand for high-speed/high-quality translation systems increased, MT became increasingly popular ~\cite{chap2.4_20}.MT has been presented via four distinct approaches: rule-based machine translation(RBMT), statistical machine translation(SMT), hybrid machine translation(HMT), and neural machine translation(NMT) ~\cite{chap2.4_21}. 
    
\end{itemize}

\subsection{Summary}

Section~2 presents the overview of the related works to the paper. First, we discussed a brief background and the type of fundamental classification of RS, including general RS unified RS, followed by the main focus on sequential RS.
 Then, we also investigate the CRS and its attributes based on the interaction, characteristics, computational tasks, knowledge and architectures. In the next step, we analysed chatbot history and chatbot-based CRS. Finally, we focused on the concepts of natural language processing and intents taxonomy in chatbots. Table 2.2 shows the prominent publications in the literature.
\section{Methodology}
This section explains the proposed model to identify user intent in CRS. The primary purpose of this model is helping to determine the user intent in human-machine interaction. One of the main contributions of this approach is to contextualize raw data and discover the importance of the selected features, which extracts powerful insights from data. The presented pipeline enhances the understanding of feature analysis and predicted results.
This system achieves higher predictive results by considering the contribution of the significant attributes and using reverse engineering and deep learning algorithms to help experts in the field of conversational artificial intelligence.

Performance evaluation is achieved based on different machine learning models and approaches. We leveraged the transformer base models for this proposed approach. Transformers ~\cite{chap3_11} by tracking the relations of sequence data help in various fields of artificial intelligence. In the last few years, transformers located a prominent position among the neural network models and received outstanding success in NLP.
The model architecture is illustrated in Figure 3.1 in different steps in a pipeline, and the following sections mention the components in detail.
The procedure includes data annotation by crowdsourcing, extracting the relevant features and using reverse feature engineering to contextualize data,  data curation (cleaning, organization, extraction and enrichment), and intent recognition algorithm to estimate the probability of the specified intentions and extracted insights.

\begin{figure}[!ht]
    \centering
    \includegraphics[angle=0, scale=0.52]{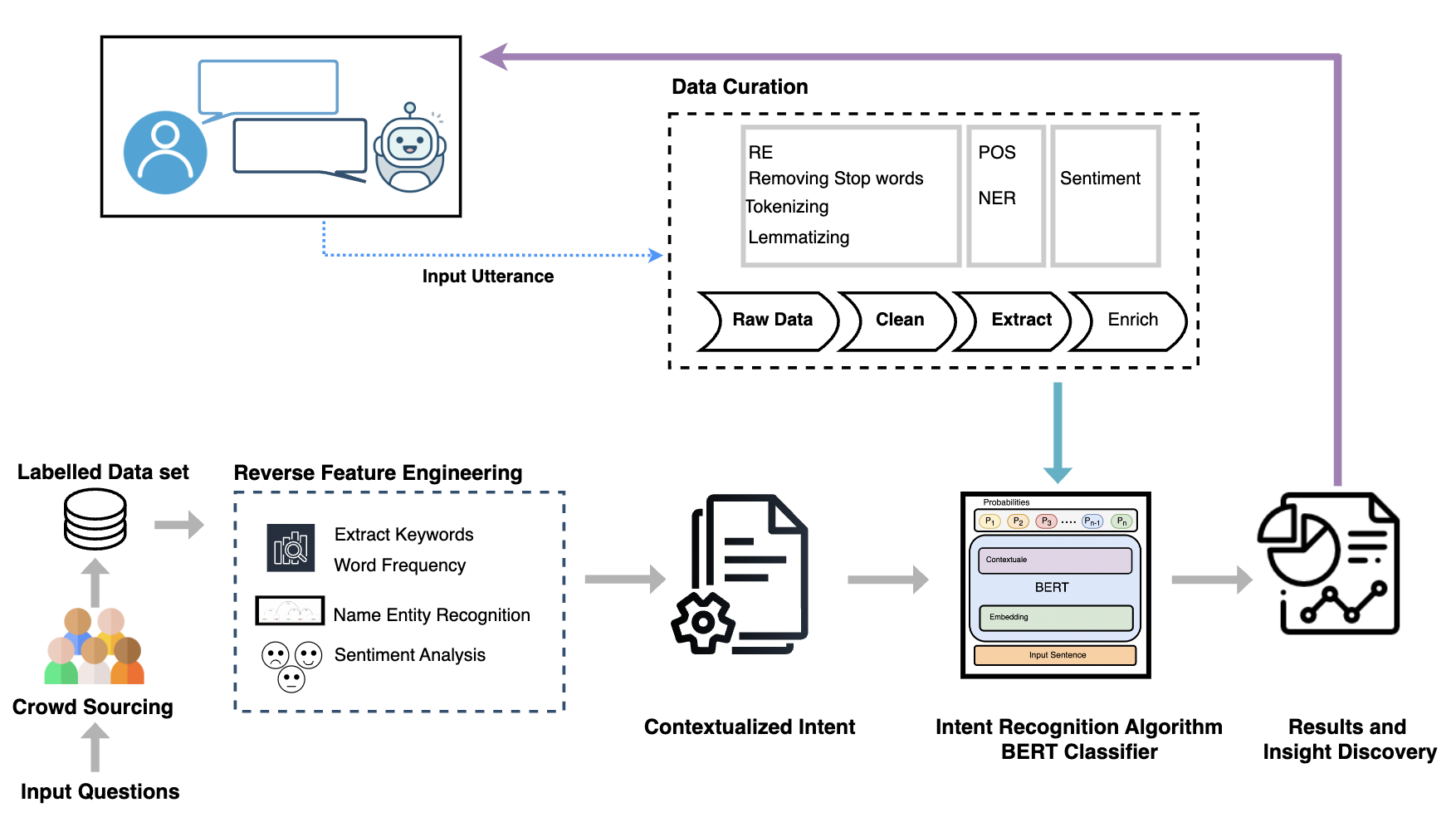}
    \caption{An overview of the proposed model.}
    \label{fig:method}
\end{figure}

\subsection{Data Annotation}
Data labelling is vital for most ML-based problems and leading restrictions for artificial intelligence projects in organizations. Data annotation regarding the sentiment and intent depends on the query divided into text, image, video, and audio. 
Data annotation is the labelling data technique to generate accurate results.
Today, crowdsourcing is considered an online marketplace where platforms like Amazon Mechanical Turk (MTurk)\footnote{https://www.mturk.com/}, Upwork\footnote{https://www.upwork.com/}, and Appen\footnote{https://appen.com/} hire the workforce for annotating a massive amount of data.
These platforms create good opportunities for industries to outsource their jobs. Ruled-based crowdsourcing can helps companies to take steps according to their policies and requirements~\cite{chap3_2}.

In the first stage of this proposed algorithm,
i) attempt to identify the required intents, i.e., the purpose of the conversational system(e.g., answer, question) with the help of the Concepts Task Designers (CTD).
ii) Perform data labelling process.
iii)Train the outcome of annotation before launching the results based on the professional designers' feedback.

\subsection{Data Curation}

Data is a vital asset for companies. Accordingly, data quality and proper approaches to extracting knowledge from them are beneficial investments. Reusing and integrating data depending on different needs and scenarios significantly impacts the quality of the results obtained for companies and organizations. 

Data curation as a multi-dimensional problem is categorized into different activities depending on the various criteria and requirements, such as scale, human-data interaction, automation, and interoperability~\cite{chap3_6}.
In this phase, we use data curation~\cite{chap3_1} to turn the input utters data into contextualized knowledge and transfer it into the algorithm.

The considered data curation pipeline consists of three separate phases: cleaning the raw data, extraction, and enriching stages.
(i) Cleaning the raw data is the procedure of creating patterns by regular expression(RE) \footnote{https://www.regular-expressions.info/tutorial.html}, removing stop words, tokenizing, and lemmatizing the input utterances.
(ii)Feature Engineering and Extraction: include extracting Part of Speech(POS), Name Entity Recognizer(NER), Verb Frequency(VF), Keyword Frequency(KF), and synonyms.
(iii)Enrichment: Data enrichment creates value from raw data, and the analytics will improve by eliciting sentiment analysis, synonyms and other enrichment techniques from them.
\medskip

\subsection{Reverse Feature Engineering }

\textbf{Feature Engineering (FE):} 
It is the group of Machine Learning (Ml) techniques used to determine, manipulate, and transform the raw input data into usable features in statistical, supervised learning, and unsupervised learning. The purpose of extracting the most related features is to facilitate and simplify the data transformations and improve the model performance.
Feature engineering includes: 
(i)Feature creation, which develops the most valuable variables based on the selected model;
(ii)Transformation, which illustrates the visual form of data by converting features from one representation to another; 
(iii)Feature extraction, without skewing the actual relationships or essential information, tries to extract features to determine valuable information;
(iv)Exploratory data analysis is a simple tool that explores properties and improves the understanding of data. 

\textbf{Reverse engineering (RE):} This is a process that collects technical data and creates a manual redesign strategy. The RE cycle enables the development of advanced computational standards and measures the efficient comparison of concurrency~\cite{chap3_3}.

\subsection{Contextualizing Data }
It provides meaningful information by integrating relevant information from various sources to simplify data interpretation and analysis of big data and metadata~\cite{chap3_4}.
This procedure facilitates decision-making by summarizing the amount of reasoning required (e.g., refine, aggregation). Therefore, it reduces unrelated data and improves the extracted knowledge from them.
Analyzing contextual data creates powerful assets for companies to gain insights from patterns and methods to achieve their purposes.
In this project, we proposed contextualizing conversational data based on the requirements to specify human intents for machines.

\subsection{Intelligent Intent Detection and Results Discovery}

According to the outcomes of reverse feature engineering and contextualized data, there are different approaches to understanding the conversation's overall meaning and intentions. Classic matching algorithms and Machine Learning matching systems are practical approaches to solving such problems.

Traditional keyword matching is a searching technique to find the specific string, keyword, or binary data through unique patterns and rules in system~\cite{chap4_1}.
Any ideal keyword set has two main properties to build meaningful outcomes: non-empty and finite groups. Meanwhile, there are other properties that we can measure the performance and complexity of the algorithm that has a significant impact on the algorithm, such as keyword set size and length ~\cite{chap4_2}. In general, user intent detection is an extensive task and is an example of text classification, where we expect the classifier learns to predict the target class.
There are three main types of  label classification:
\bigskip

\textbf{\emph{(i) binary classification}}: is a supervised ML problem where data classify into exclusive categories in a labelled dataset, for instance, zero or one, true or false. This type of classification is used in different scenarios like fraud or spam detection and medical diagnosis. Binary classification is a simple way of image prediction (e.g., spam or non-spam detection).
\medskip

\textbf{\emph{(ii) Multi-class classification}}: is also a supervised ML- a problem that is trained on a dataset with more than three categories or classes. Multi-class classification is divided into single-label and multi-label categories and creates the last category. Unlike single-label, multi-label classification does not have a limit on the number of classes.
\medskip

\textbf{\emph{(iii) Multi-label classification}}: similar to previous approaches, is a supervised ML- algorithm. Multi-labels are divided into zero or more labels for each sample. In a conversational turn, for instance, greeting and asking questions are more than one intent.
\bigskip

Many conventional machine learning approaches, including the
k-nearest neighbours, and decision trees, use to solve the multi-class problem. Also, deep learning methods, for instance, Recurrent Neural Networks (RNNs) or Transformer-based algorithms, work to learn by relying on textual feature types and discrete feature representations.
In our case, we focus on natural language tasks through a transformer-based algorithm to classify the user's intent with a well-known technique such as Bidirectional Encoder Representation of Transformers (BERT).

A \textbf{transformer} is a new deep-learning approach mainly used in NLP and Image processing. Transformers are based on the self-attention mechanism proposed in 2017~\cite{chap3_12}. Transformers with the sequence-to-sequence architecture replaced the recurrent neural networks by processing the complete data input all at once and reached considerable efficiency improvements.

\begin{figure}[!ht]
    \centering
    \includegraphics[angle=0, scale=0.4]{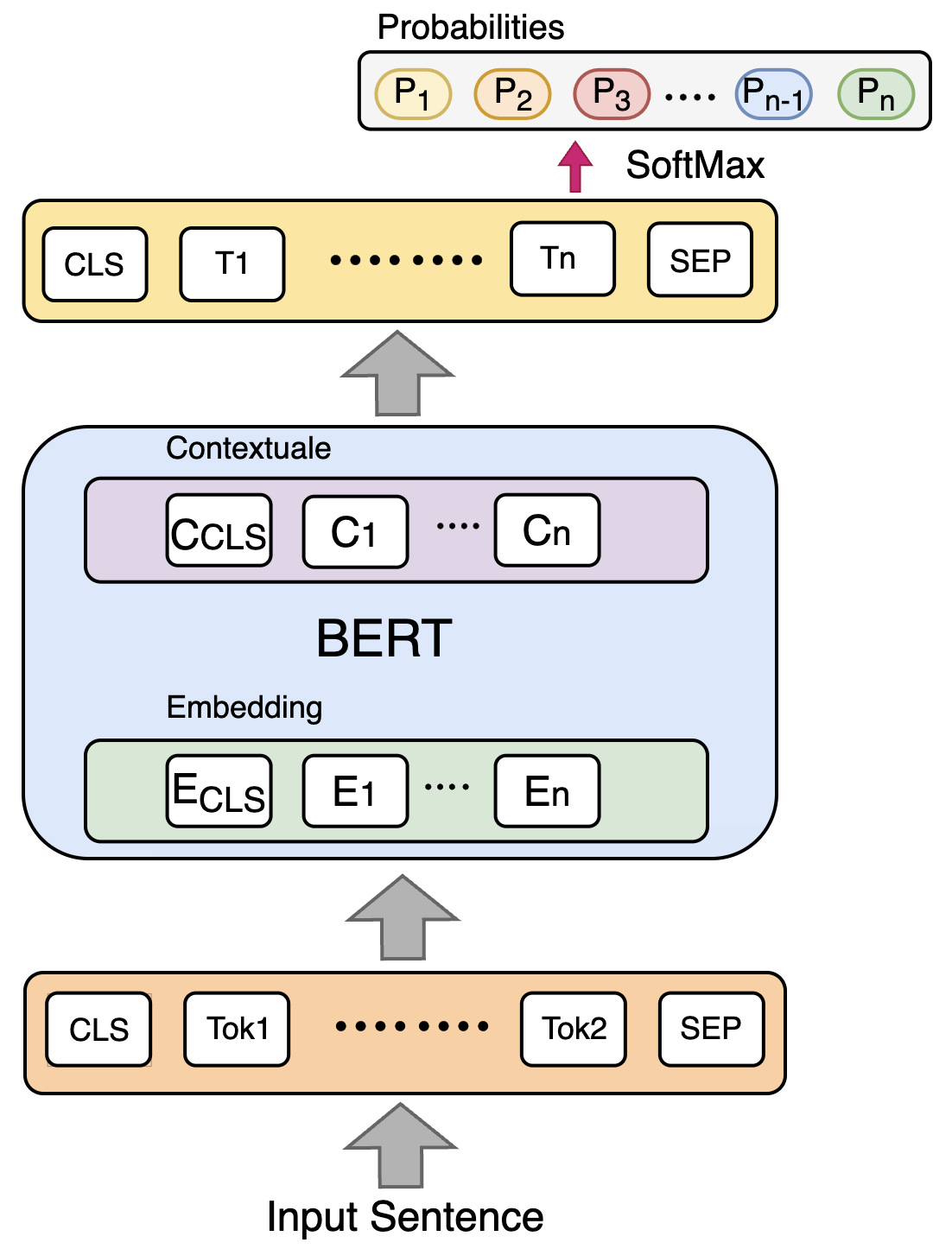}
    \caption{BERT architecture for multi-label classifications~\cite{chap3_7}.}
    \label{fig: BERTArchitecture}
\end{figure}

\textbf{BERT} was proposed in 2018 by Google as a powerful language-based model to apply to lots of NLP tasks~\cite{chap3_7}. This approach covers the context-free(e.g.,word2vec to present each word in vocabulary as a single-word embedding) and contextual models. The contextualized models represent words based on the other words and the position in the sentence. Google Search is a famous example of a system that uses BERT. 
This model uses attention-based approaches instead of the sequential nature of Recurrent Neural Networks(RNN).

BERT is the state-of-the-art model in many sequence-based language understanding tasks, such as generating conversational responses and summarization. However, it helps to improve the proficiency of content understanding in various languages.
This open-source model has a significant impact on various classification problems and provides the occasion to design different classifiers such as the bioBERT (a model for biomedical content mining)~\cite{chap3_8},  SciBERT( model toward scientific content)~\cite{chap3_9}, and  TinyBERT(smaller and faster version of BERT transformer to improve performance)~\cite{chap3_10}.

Therefore, by leveraging this model, we use the multi-label classification to indicate the intent category.
We created a probability distribution for the predicted tokens using the Softmax function. In contrast to simple answers, SoftMax enables us to respond to classification queries with probability. Finally, we assign the highest probability of all the intent classes. The mathematical expression for SoftMax is shown in the following formula:

 \[S(x_i) =  \frac{e^x_i}{ \sum_{i=1}^{n} e^x_i} \]

In this proposed approach, we tried to facilitate and clarify the user's goal to help chatbot-base RS communicate more easily with people and provide a step further for organizations with an opportunity to automate communication with their customers.


\section{Experiments and Evaluation}
In this section, we present the dimensions for evaluation results of the proposed model in Section~3 on a real-world dataset and analyse how the proposed model helps to understand user intent in CRS.
First, we discuss the motivating conversational scenario to explain our approach. 
Next, we provide detailed information about real-world data.
Then we clarify experimental setups, progress and outcomes and finally explain how to evaluate the results.

\subsection{Motivating Scenario}

Intelligent chatbots help answer human inquiries in various fields (e.g., health ~\cite{chap4_5} and education~\cite{chap4_6}) and understand the user's intent, drive to provide accurate responses quickly. This opportunity empowers users to express their requirements, and intelligent communication effectively improves their social journey. As a motivating scenario, we focus on detecting the user intent in chatbot base CRS to help organizations use the advantages of market needs by discovering the purpose of users from different aspects. Therefore, we target three other challenges that organizations face in them as follows:

\textbf{Use Case 1: Transferring the knowledge generated during the human-chatbot interaction to business analysts.}

Identifying and analyzing the problem to choose the correct actions is an undeniable effort for organizations. The ability to diagnose a problem and implement a solution as quickly as possible is a vital asset for companies. Therefore, transferring the result of human-machine interaction to experts accelerates problem-solving and provides an excellent opportunity for professionals to prepare systematic structures for troubleshooting. Since it requires knowledge and a logical approach as a valuable asset to finding the solution, identifying the causes of problems will accelerate and improve system efficiency.

\textbf{Use Case 2: Improving processes, e.g., system proficiency and time reduction.}

In the past decade, having dialogue assistance to provide customized recommendations has become more common among companies.
These dialogue systems are created to implement specific objectives for decisions making based on information filtering. Conversational AI is making rapid progress in various fields. Dealing with customer inquiries is stressful and challenging, for example, in the banking system. Hence, understanding user concerns and problems help companies to have a better view of market needs. In addition, it assists companies boosts and presenting the underlying infrastructure. Finally, expand the conditions for the production of an automation system.
Compared to a traditional system or a human agent, automation provides users with more efficient and desirable services.

\subsection{Dataset}
High-quality data is considered a valuable asset for the improvement and development of the industry in various branches. Real-world data provides irrefutable evidence to reduce high investment risks and provide greater flexibility, predictability, transparency and certainty to drive robust and consistent decision-making.
We used the MSDialog dataset, which includes over 30,000 high-quality labelled question-answering dialogues.

MSDialog is based on conversations with the technical Microsoft forums from different products such as Windows, Bing, Skype, and Office from 2005 until 2017. These conversations are interactions between the information seekers (user-generated questions) and high-quality responses from Microsoft Community technical support and Microsoft Most Valuable Professionals(MMVP).
MSDialog-Complete is the complete set of conversations. This dataset includes different interpretation subsets such as Dialog-Intent and MS Dialog-Response Rank.

In this paper, we used the Dialog-Intent dataset. The conversations in this dataset are annotated via crowd-sourcing through the Amazon Mechanical Turk(MTurk)\footnote{https://www.mturk.com/}. The MSDialog includes about 2,000 multi-turn information-seeking dialogues with 10,000 utterances that the selected subset has the specific criteria: \emph{i)}the 2199 conversations with 3 to 10 turns\emph{ii)} between 2 to 4 participants, \emph{iii)} consists of at least one correct response.

The dataset contains information regarding the conversation media, details around dialogue such as time and date, frequency of the conversation, title, and more elements for each utterance.
In addition, each utterance contains the user's unique id, specified types of the actor as a user or agent, the priority of the sentences in the conversation, the type of each sentence (is it a response utter or not), and how the answer is helpful for the other users and the intent tags. Part of the conversation's utterances and details are represented as follows in Figure 4.1.

\begin{figure}[!ht]
    \centering
    \includegraphics[angle=0, scale=0.45]{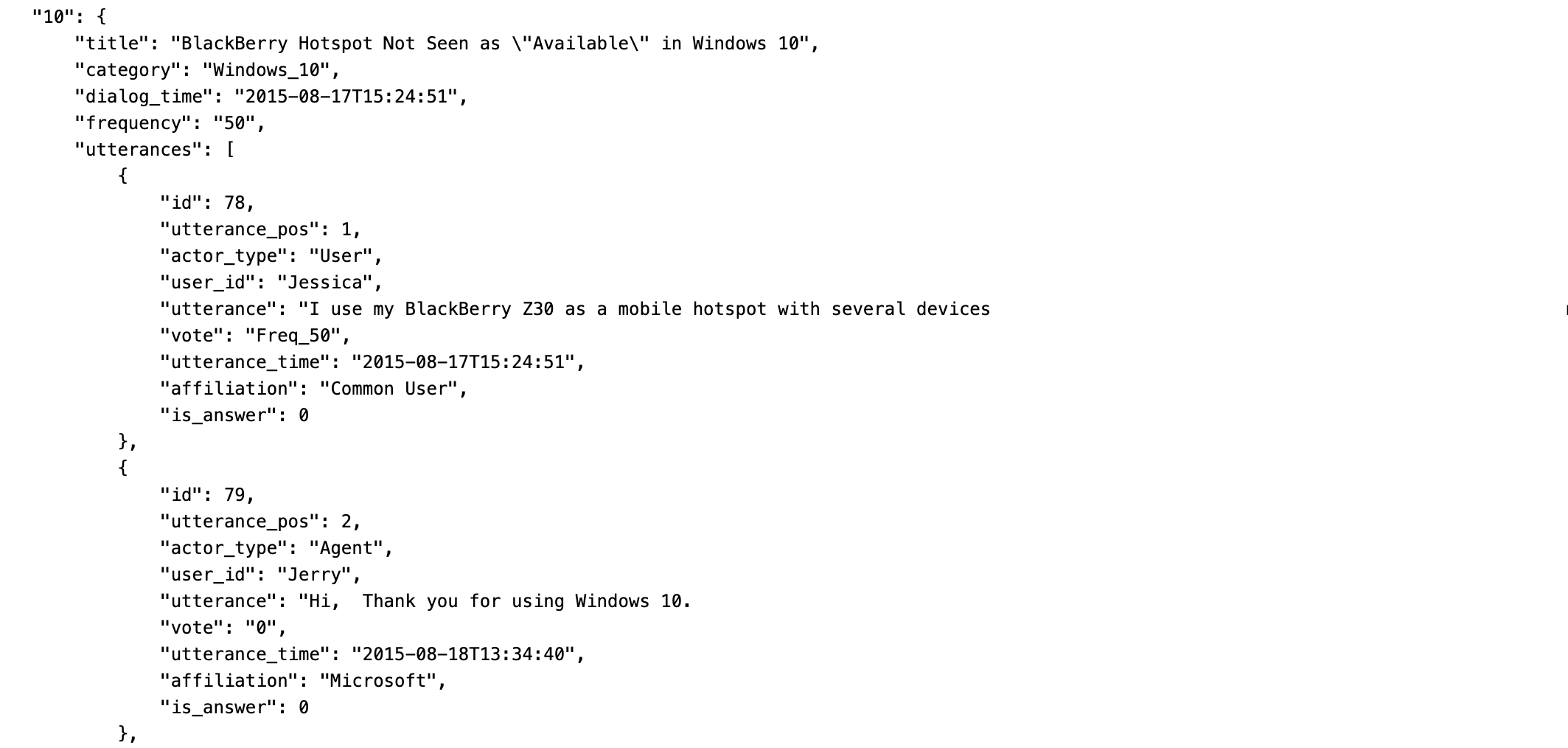}
    \caption{Example of preliminary format for the conversations and utterances in MSDialog Dataset that used to extract conversations and labelled for the analysis.}
    \label{fig:conv}
\end{figure}

\subsection{Experimental Setting}
All the experiments are run on a machine with Apple M1 Pro and 16GB RAM. The experiments are performed based on Python programming  language and used various open-source software libraries, visualization libraries and platforms, as follows:
PyTorch,  Hugging Face Transformers, NLTK, SpaCy, Scikit-Learn, Pandas, Numpy, Seaborn, and Matplotlib.

\subsection{Experimental Results}

\subsubsection{Preprocessing data}
Evoking the main attributes from a corpus is the first challenge. 
In this stage, we perform pre-processing to extract the utterances with all related details from the conversations in the raw dataset. Then, we use elements like actor type (e.g., user or agent), utterances, and intents.  

In this unique dataset, by analysing the utterances in the first step, we attempted to remove the unnecessary features by performing \emph{(i) Regular expression} to create a pattern for searching in texts, \emph{(ii) Tokenizing} to understand paragraphs by separating and breaking sentences into smaller sections as words, \emph{(iii) Removing stopwords} for vacating the most common words across a corpus, and \emph{(iv) Lemmatizing} to have a clear idea about the root (lemma) of each word. Finally, the result is a clear list of words for each utterance. Table 4.1 shows the first rows of selected features in the extracted dataset. 

\begin{table}[!ht]
        \centering
        \includegraphics[scale=0.3]{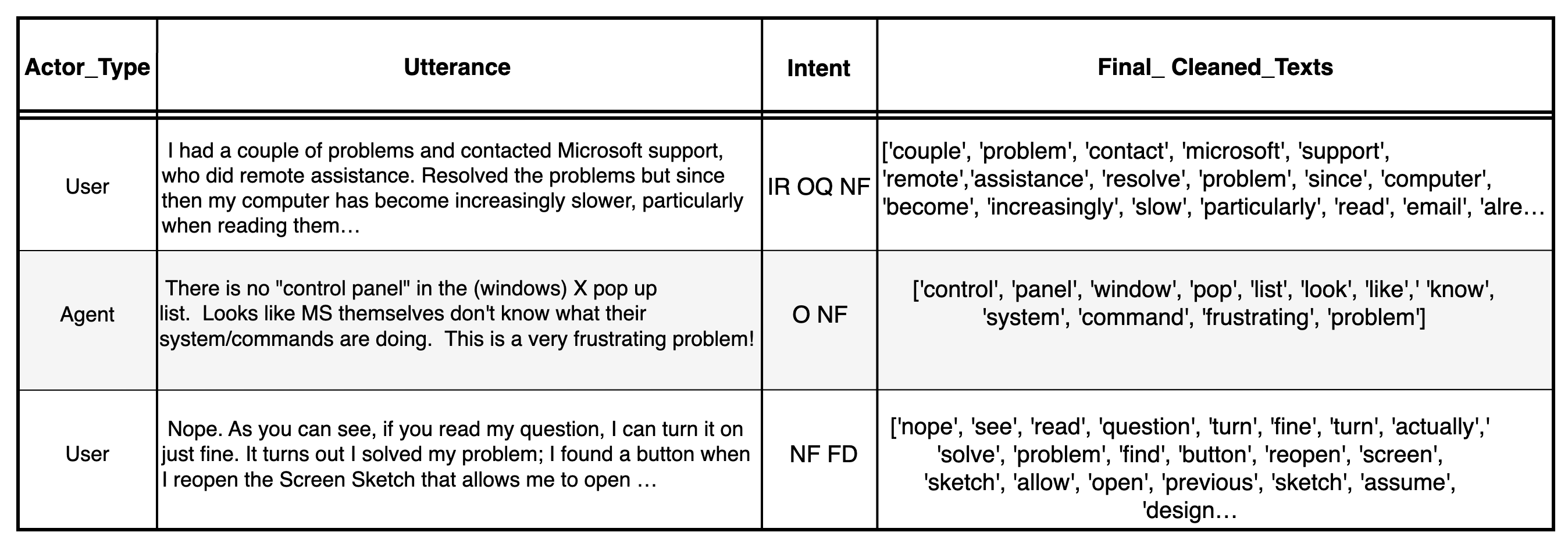}
        \caption{Snapshot of the first rows of the extracted dataset from MSDialog, regarding the essential features for understanding the intent, which includes the type of the actors (user or agent), extracted utterances from conversations, labelled intents, and the cleaned/preprocessed utters.}
        \label{tab:dataset}
    \end{table}
Best results and increasing the accuracy of the models require high-quality data. Also, efficient data labelling is increasingly critical in classification problems.
Data annotation is classifying and labelling data to make decisions and take appropriate actions. high-quality text-annotation
contain intent, sentiment, and semantic annotations. MSDialoug annotated by 12 types of intents that Figure 4.2 illustrates these different types of intents.

\begin{figure}[!ht]
    \centering
    \includegraphics[angle=0, scale=0.28]{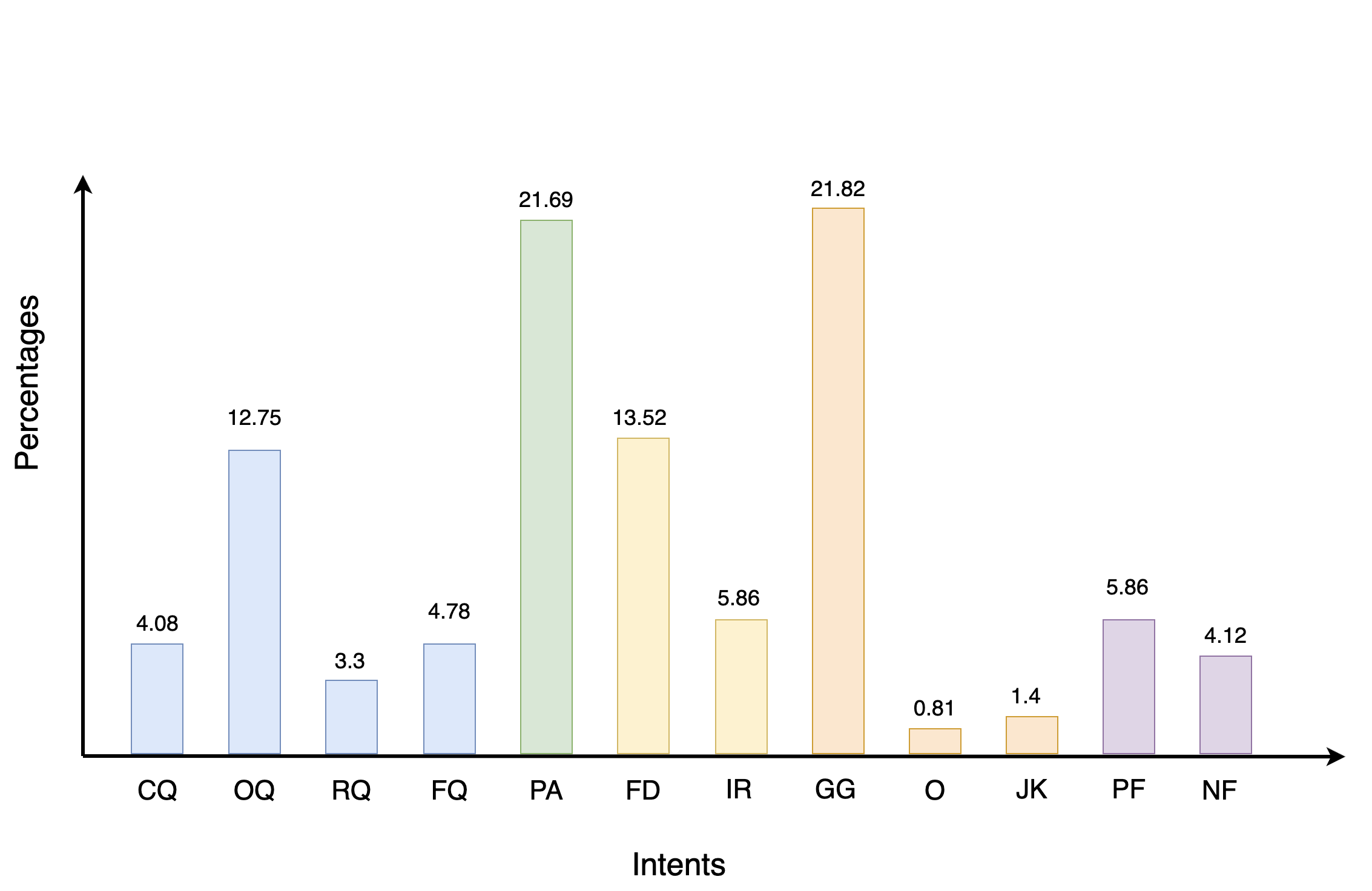}
    \caption{Intent labelled distribution of utterances in the MS Dialog-based dataset.}
    \label{fig:label}
\end{figure}
\bigskip

Forecasting the overall purpose of the human is beneficial for various applications and scenarios. As seen in the figure above, user intention distribution is simultaneous in some utterances. For the initial effort, we extract intent collusion and learn more about distributions. We integrate the conversation goals into five categories: questions, answers, feedback, further information, and no information. The results are shown in Figure 4.3.

\begin{figure}[!ht]
    \centering
    \includegraphics[angle=0, scale=0.31]{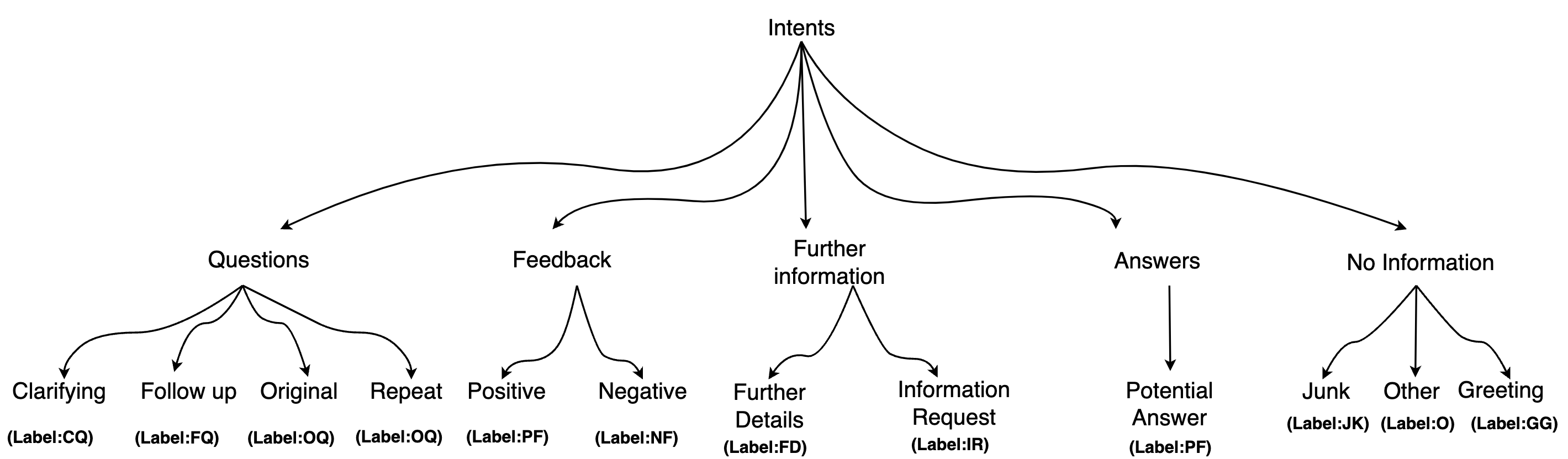}
    \caption{Intent classifications based on the proposed method and specified labels.}
    \label{fig:Intentclassifications}
\end{figure}

\subsection{Feature Engineering and Extraction}

In the next stage, we attempted to extract the pre-analysis of the dataset in-depth in a couple of steps:

\hspace{0.3cm}

\textbf{\emph{i) Sentiments:}} Understanding user sentiment is a critical element of business, and the opportunity to follow user conversations with chatbots helps companies gain qualified market insights.
Sentiment analysis collects information about how people speak and believe about a specific industry.
Analyzing user sentiments from paragraphs or sentences of a corpus helps to understand the polarity of a text. This polarity is broken down into different aspects, including specific emotions, feelings, and intentions.
This research extracts non-graded sentiment analysis divided into positive, negative, and neutral.
Figure 4.4.a shows sentiment analysis for variously defined intents, and Figure 4.4.b illustrates the overall sentiment analysis of the dataset without assuming the specific intention.

\begin{figure}[!ht]
    \centering
    \includegraphics[angle=0, scale=0.3]{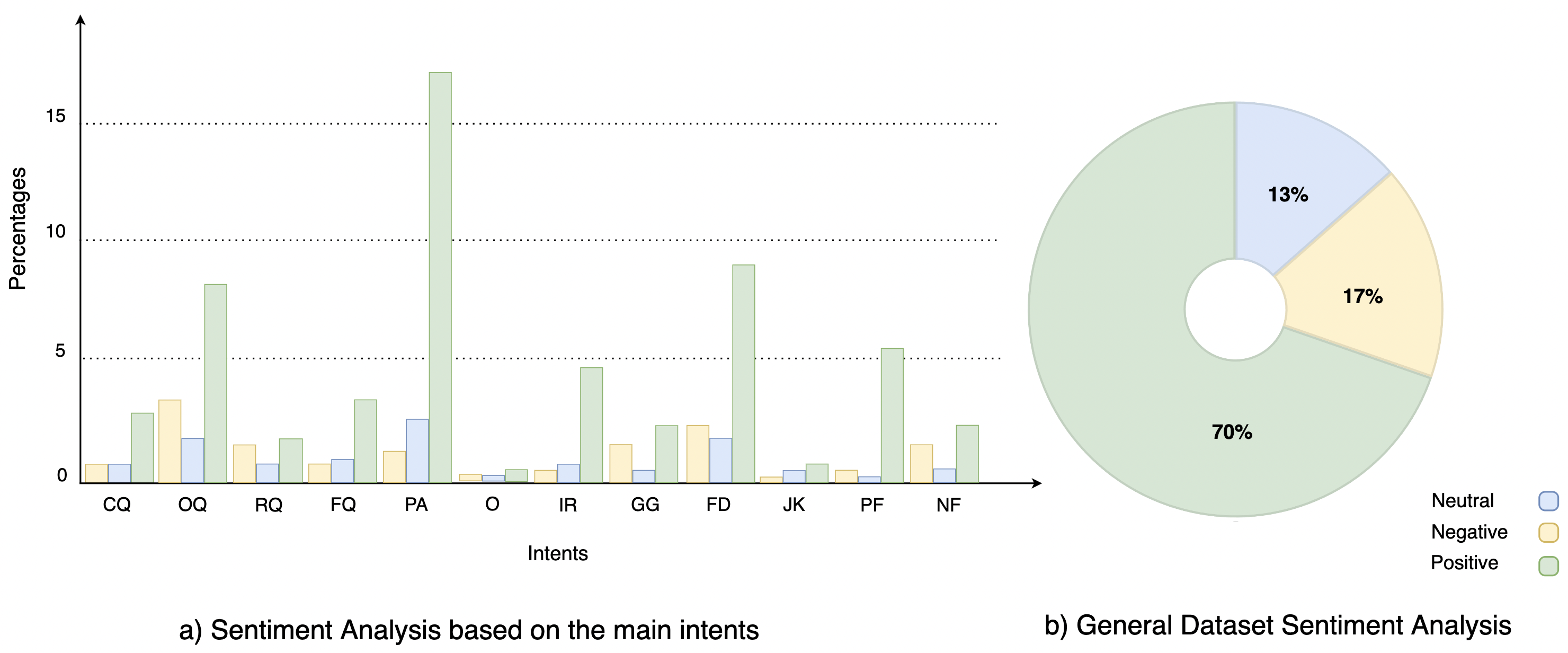}
    \caption{Overall and defined intents sentiment analysis in the dataset.}
    \label{fig:Sentiment}
\end{figure}
\vspace{-\baselineskip} 
\medskip

\textbf{\emph{ii) Name Entity Recognizer (NER):}} 
An entity is any real-world object that contains a term or a set of words.
Therefore, a person's name, geographic location, and products are included as a named entity. NER is the process of determinating named entities in text, depending on the defined categories. 
Classification of contents(provide insights around the text), recommendation and search engines( improve the speed and quality of search), and helping the customer and human resources (decrease the workflow and response time) are good use cases of NER. We leverage NER to maintain a high-level overview and understanding of this large dataset. NER extraction can be performed relying on rule-based or ML approaches. Figure 4.5 is an example of NER on the MSDialog dataset via the spaCy library.

\begin{figure}[!ht]
    \centering
    \includegraphics[angle=0, scale=0.25]{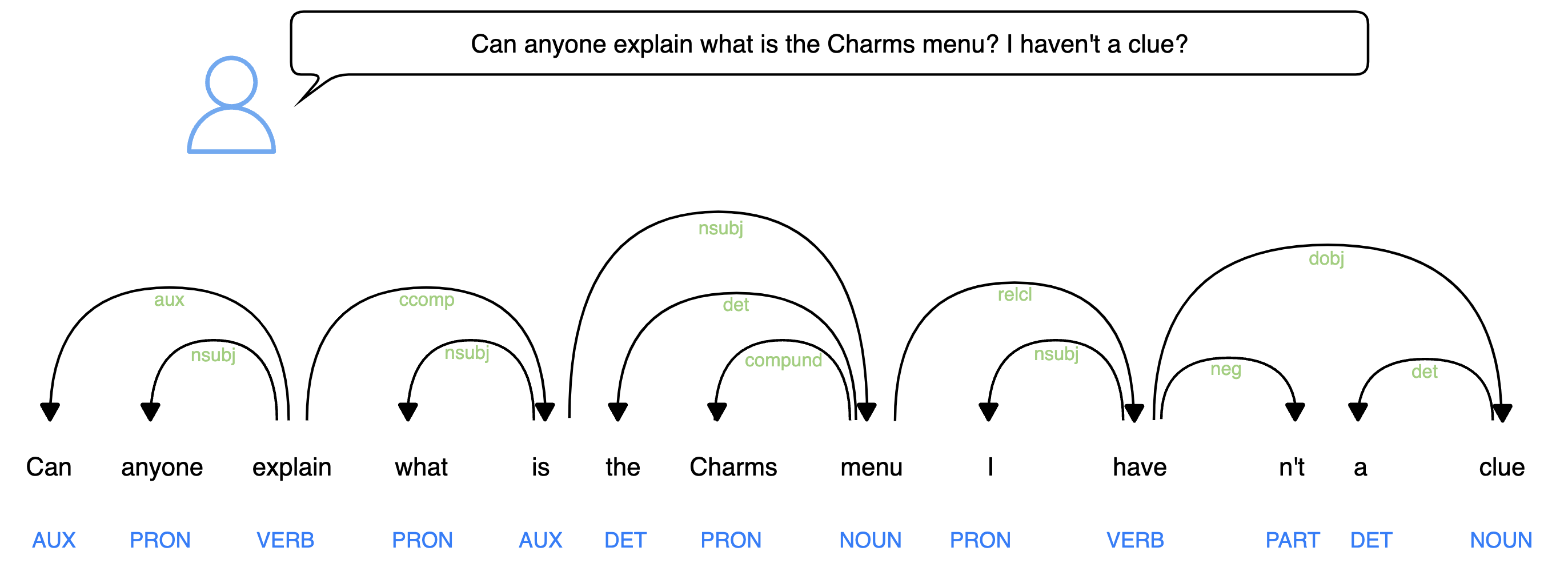}
    \caption{Example of NER Based on the dataset.}
    \label{fig:NERBased}
\end{figure}
\bigskip

\hspace{0.5cm}\textbf{\emph{iii) Word Frequency(WF):}}
Depicts the occurrence of frequent words in a text document. A high quantity of WF for each word represents the influence and importance of each word in a sentence and helps to highlight the most critical terms in a corpus. At this stage, we have examined and analysed the most frequent words in the utterances based on the defined primary intents question, answer, feedback, further information, and no information. Figure 4.6  illustrates the words with the most frequency in each utterance.
\begin{figure}[!ht]
    \centering
    \includegraphics[angle=0, scale=0.28]{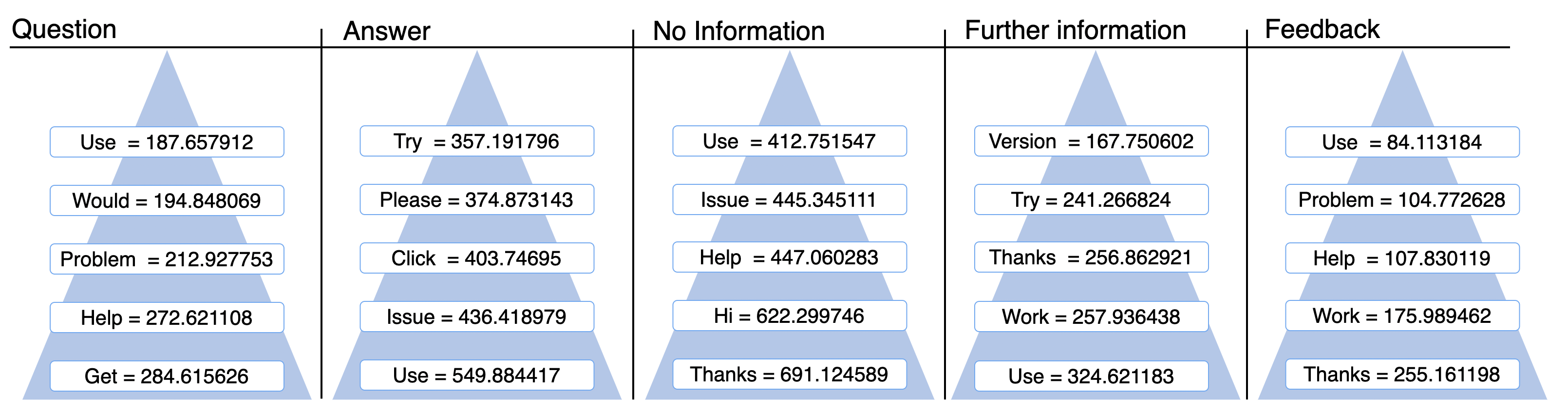}
    \caption{Words with the highest frequency in the dataset.}
    \label{fig:WF}
\end{figure}
\subsubsection{Contextualizing Data}
To understand the meaning of the context, we interpret the extracted data with different approaches and methods. Collecting all the relevant information about the specific objective highlights the importance of contextualized knowledge. Data contextualizing is an integral approach to connecting extracted data to domain knowledge and providing a broad understanding of the objectives identified in the dataset. Table 4.2 shows part of the contextualized intents.

\begin{table}[!ht]
\centering
\includegraphics[scale=0.3]{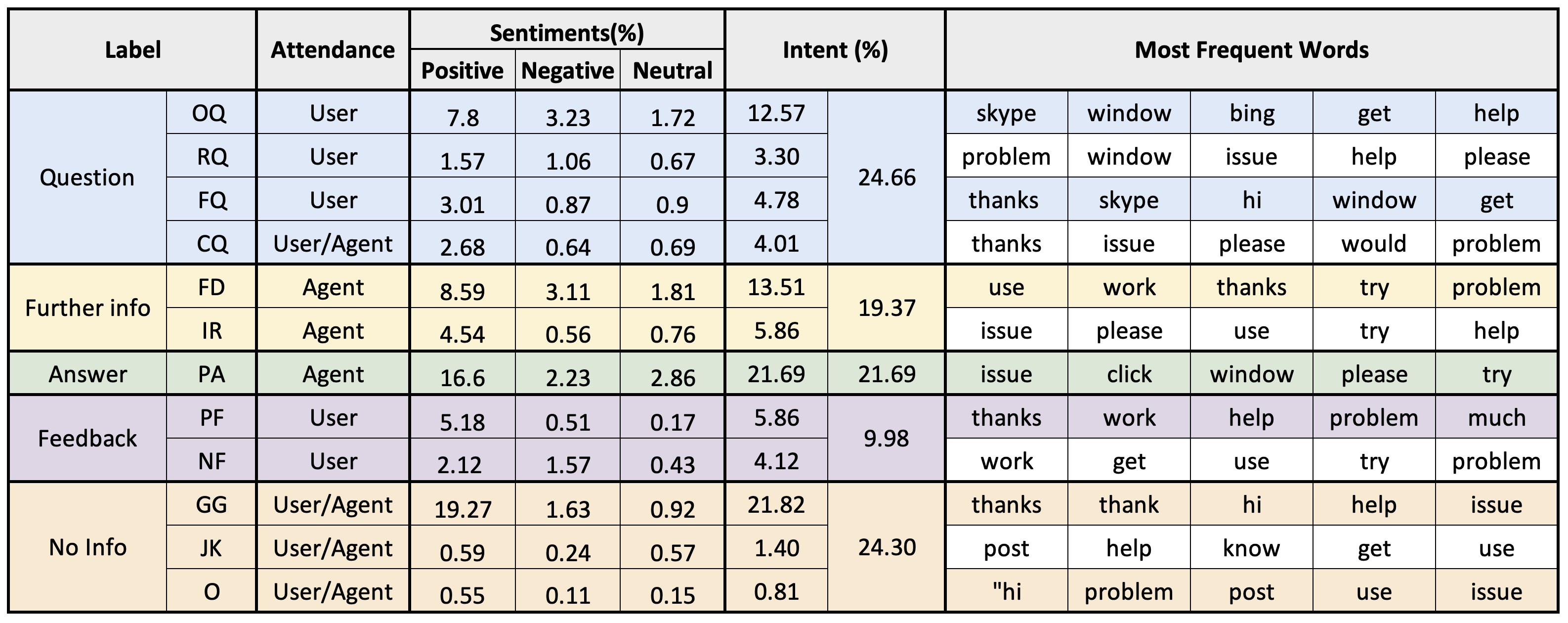}
\caption{Part of contextualized intent by considering the distribution of labels, sentiments, and word frequency.}
\label{tab: Contextualizeddata}
\end{table}

\bigskip

\subsection{Evaluation}

\subsubsection{Evaluation Objectives} 
    
    To evaluate the correctness of our approach, we performed a study experiment and attempted to validate the following hypotheses:
 \medskip

   \textbf{H1:} Recognizing customer intents throughout a chatbot may improve user experience.\hfill 
   
\hspace{1cm} a) To receive appropriate responses quickly.\hfill

\hspace{1cm} b) To obtain more accurate and effective responses to their requests.

   \textbf{H2:} Autonomous intent detection may assist domain experts and analysts.\hfill
   
\hspace{1cm} a) To scale up services to a higher number of customers.  \hfill 

\hspace{1cm} b) To improve customer service quality by identifying problems and challenges.

\subsubsection{Evaluation Setup}

\begin{itemize}
    \item \textbf{Evaluation of H1:} This hypothesis assumes that recognizing customer intents throughout a chatbot may improve user experience to receive more accurate and quick responses. To evaluate this hypothesis, by considering a multi-label with a multi-class problem, we implement the BERT-Base Classifier and attempt to tune that. In the next step, based on the contextualized data, we implement the model to understand the relationship between the sentiment and labelled intents. Finally, by considering the user's feedback, we implement the model to understand the relationship between the user's sentiment and labelled intents based on contextualized intent. Hence we classified the model depending on the extracted sentiment "positive", "negative", and "neutral" to analyze the accuracy of feedback.
    
    \item \textbf{Evaluation of H2:} The experiment was completed in a controlled environment considering particular qualities to study our approach.
We introduced the technical viewpoints around the motivation scenario in the first step. In the next step, we discussed the model strategy with the participants. The study was performed with 12 participants in Sydney, Australia and Tehran, Iran. Participants choose from data analysis lab students with AI/ML backgrounds and industry experts focused on customer service. Therefore, most participants had technical experience in computing.
\end{itemize}

\subsubsection{Evaluation of H1: Analysis and Results}

\begin{itemize}
    \item \textbf{Hyper-parameters and BERT Implementation Details.}
\medskip

We implement the Bert Classifier via PyTorch\footnote{ https://pytorch.org/} and use the BERT-Base model (Uncased)~\cite{chap4_3} and attempt to tune that. 
In training deep learning models, hyperparameters are measurable and numerical factors that manage the learning process in the model. Hyperparameters are crucial elements in maintaining the algorithm, and they have a significant impact on the model performances.

Therefore, after training the model with the most valuable parameters in the literature, we set the maximum sequence length of 384, with a batch size of 16. Meanwhile, for training this model, we use the Adam weight decay optimizer with an initial learning rate of 5e-5 and an initial epsilon value of 1e-8 in 5 epochs. We measure the elapsed time, reset tracking variables at the beginning of each epoch, and check the loss values and time elapsed for batches. Meanwhile, we calculate the average loss over the entire training data. Table 4.3 illustrates the details of the last epoch.
\begin{table}[!ht]
        \centering
        \includegraphics[scale=0.2]{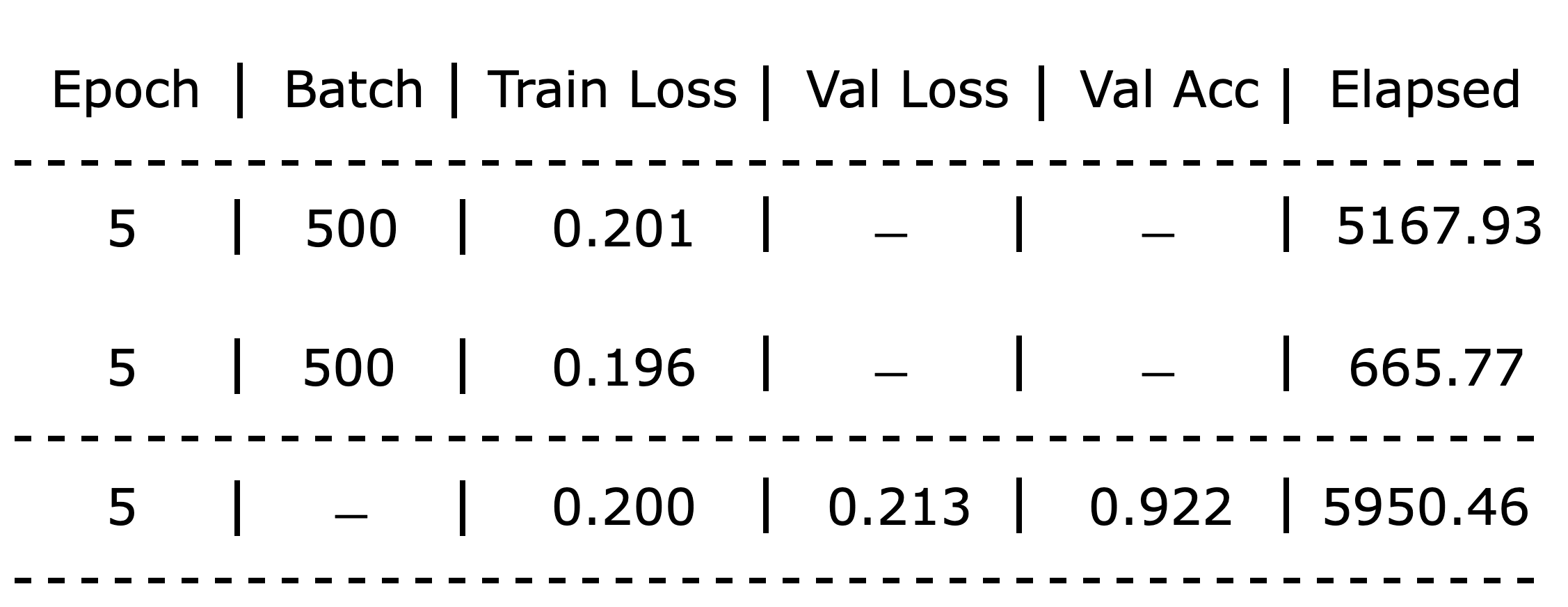}
        \caption{Running Multi-label, multi-class classifier by our tuned BERT model and the results for the last epoch.}
        \label{tab: Bert01}
    \end{table}

 \begin{figure}[!ht]
    \centering
    \includegraphics[angle=0, scale=0.3]{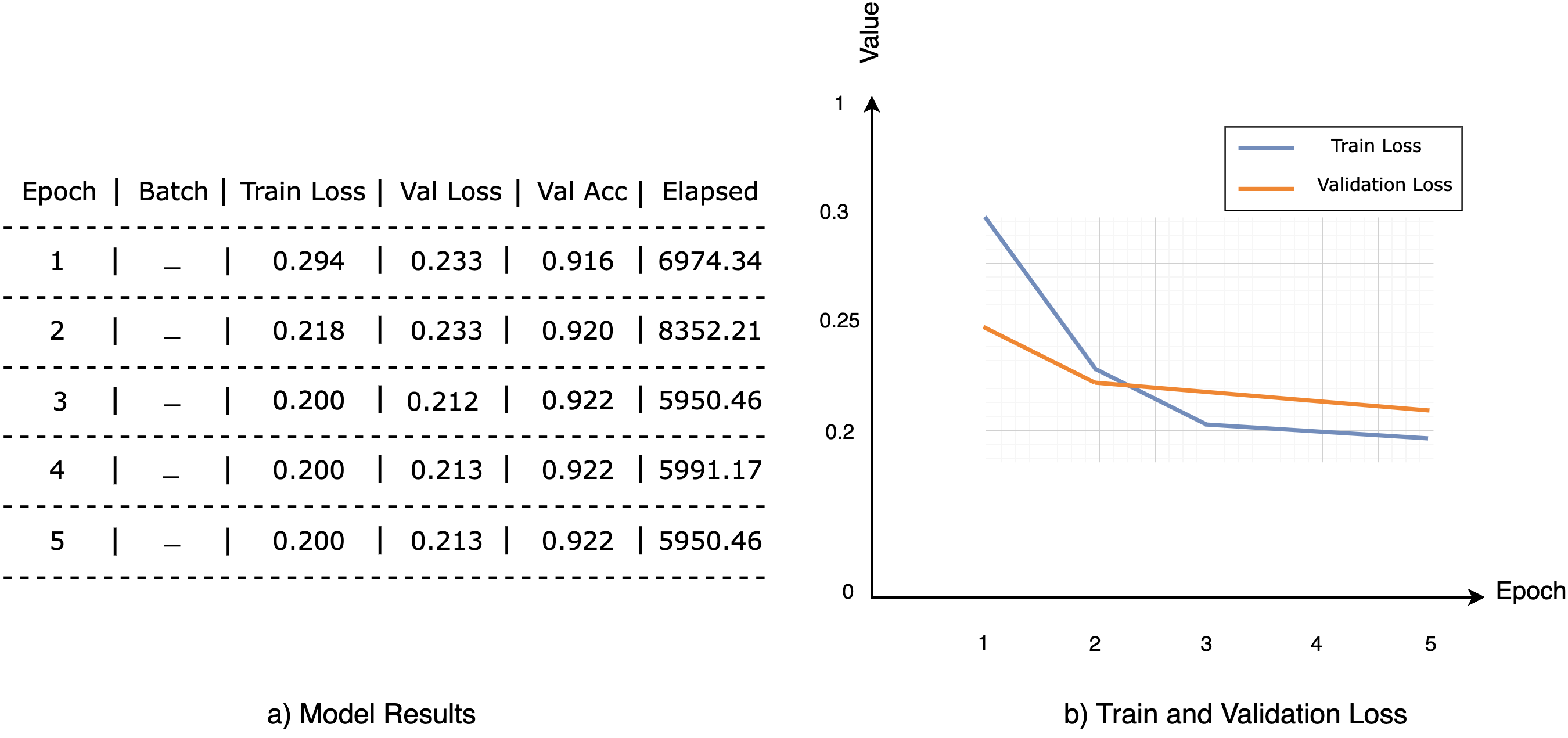}
    \caption{Analysing the Bert Results in the tuned model.}
    \label{fig:Bert02}
\end{figure}
In continuation of the analysis, Figure 4.7.a shows the results expressions for these measures. A loss value indicates the model's prediction quality, and how it is closer to zero shows the better model's prediction. Calculating the loss value on the train and validation set is to discover weights and biases in the model.
\bigskip

Figure 4.7.b shows that the loss for our model's train and validation set is not more than 0.2. Finally, we used holdout data set with a different conversation. This test dataset includes 1000 dialogues to test our model and provide an unbiased evaluation of the model.
 
\item \textbf{  Analysing the importance of the contextualised features }

In the next step, based on the contextualized data, we implement the model to understand the relationship between the sentiment and labelled intents. Hence we classified the model depending on the extracted sentiment "positive", "negative", and "neutral" to analyze the accuracy of labels on each sentiment. Table 4.4 illustrated the last epoch's details, including train loss, validation loss, validation accuracy and elapsed, for each sentiment.
\medskip

\begin{table}[!ht]
        \centering
        \includegraphics[scale=0.18]{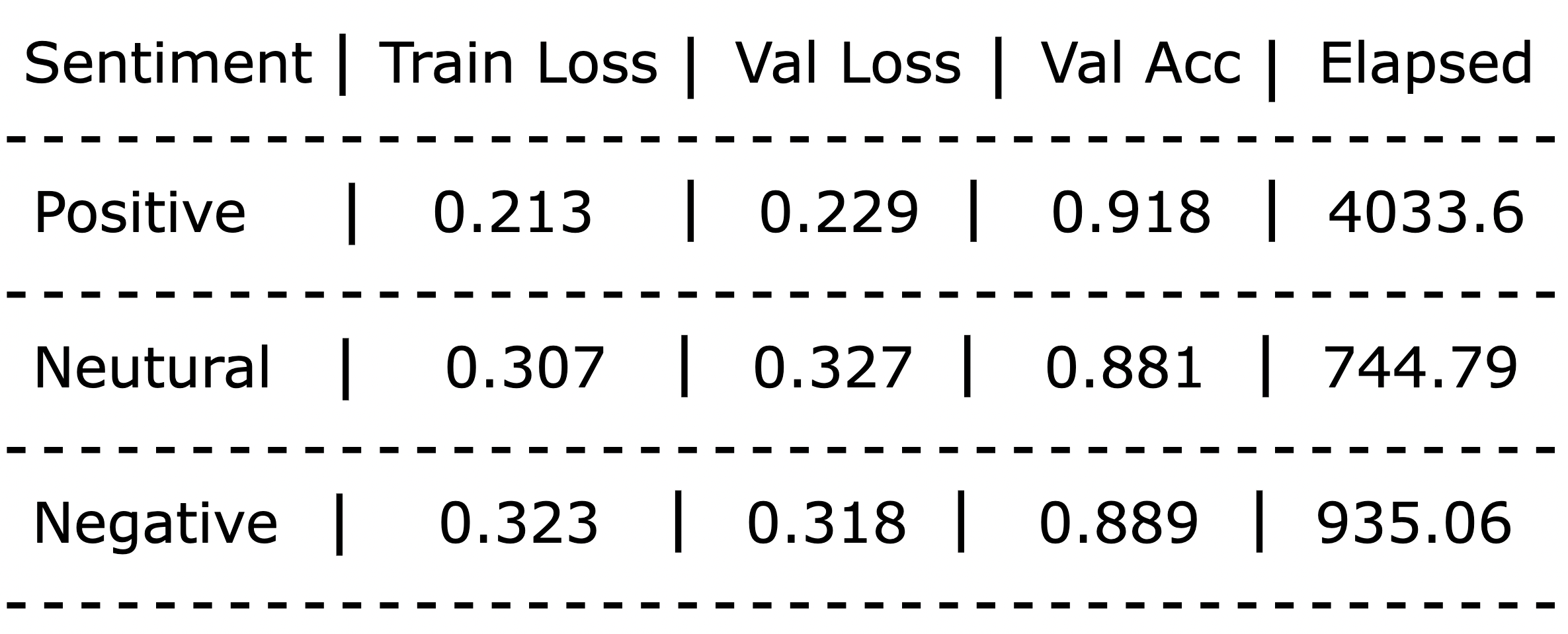}
        \caption{Running Multi-label, Multi-class classifier by our tuned BERT model to analyze the relationship between the sentiment and labelled intents.}
        \label{tab: Bert03}
    \end{table}

\begin{figure}[!ht]
    \centering
    \includegraphics[angle=0, scale=0.39]{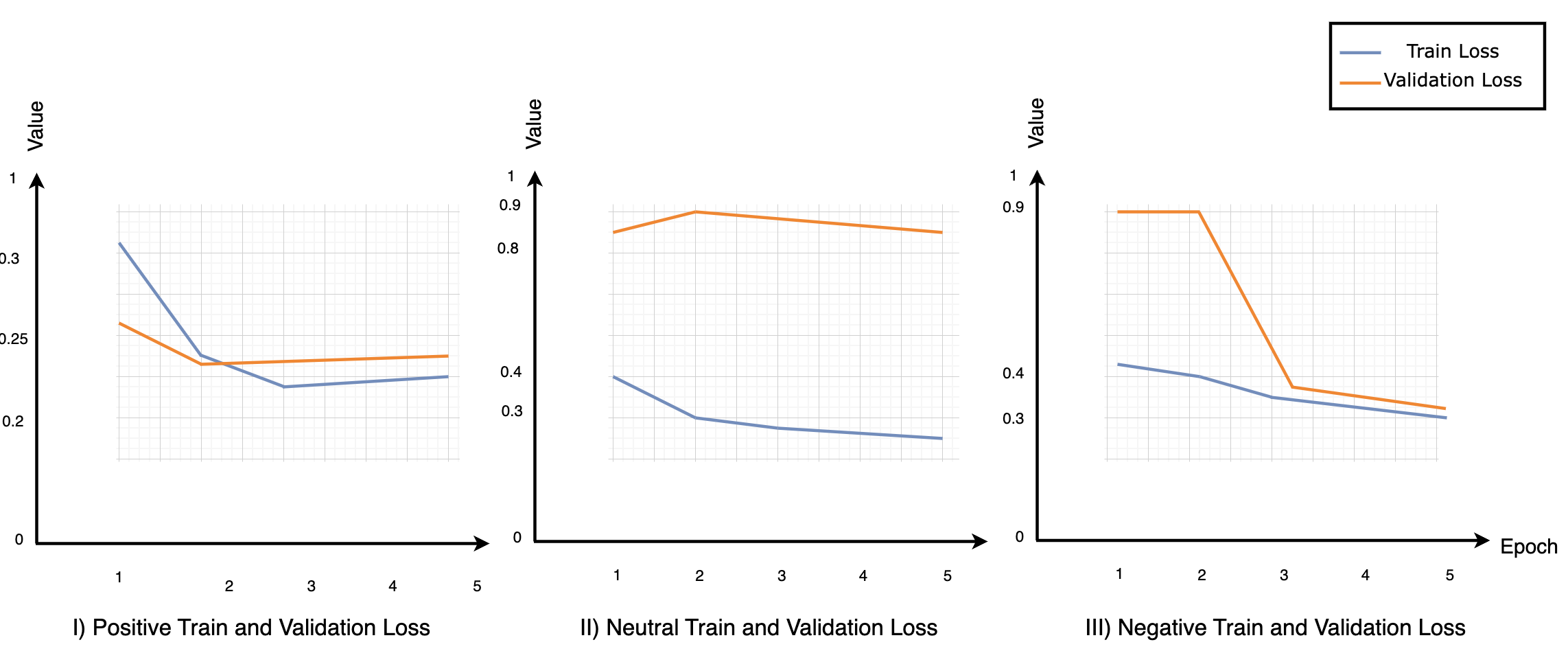}
    \caption{Comparing the loss value on the train and validation set for positive, negative and neutral sentiments.}
    \label{fig:Bert04}
\end{figure}
Finally, we compare the loss value on the train and validation set for each sentiment in the following figure.
Based on the definition of the loss value and the result of Figure 4.8, although the loss value has a decreasing direction, our model measures well on positive sentiments. 
The number of negative and neutral dialogues and these high values for loss shows that it is not the most critical inference we can learn from them.    
\medskip

\item\textbf{Analysing sentiments based on user feedback}     

We implement the model to extract the user sentiments based on their positive and negative feedback. Based on the feedback, the result shows about 89.41 per cent accuracy of our model. The result shows an accurate and quick understanding of the user's questions, providing more satisfaction to users.      
      
\begin{figure}[!ht]
    \centering
    \includegraphics[angle=0, scale=0.43]{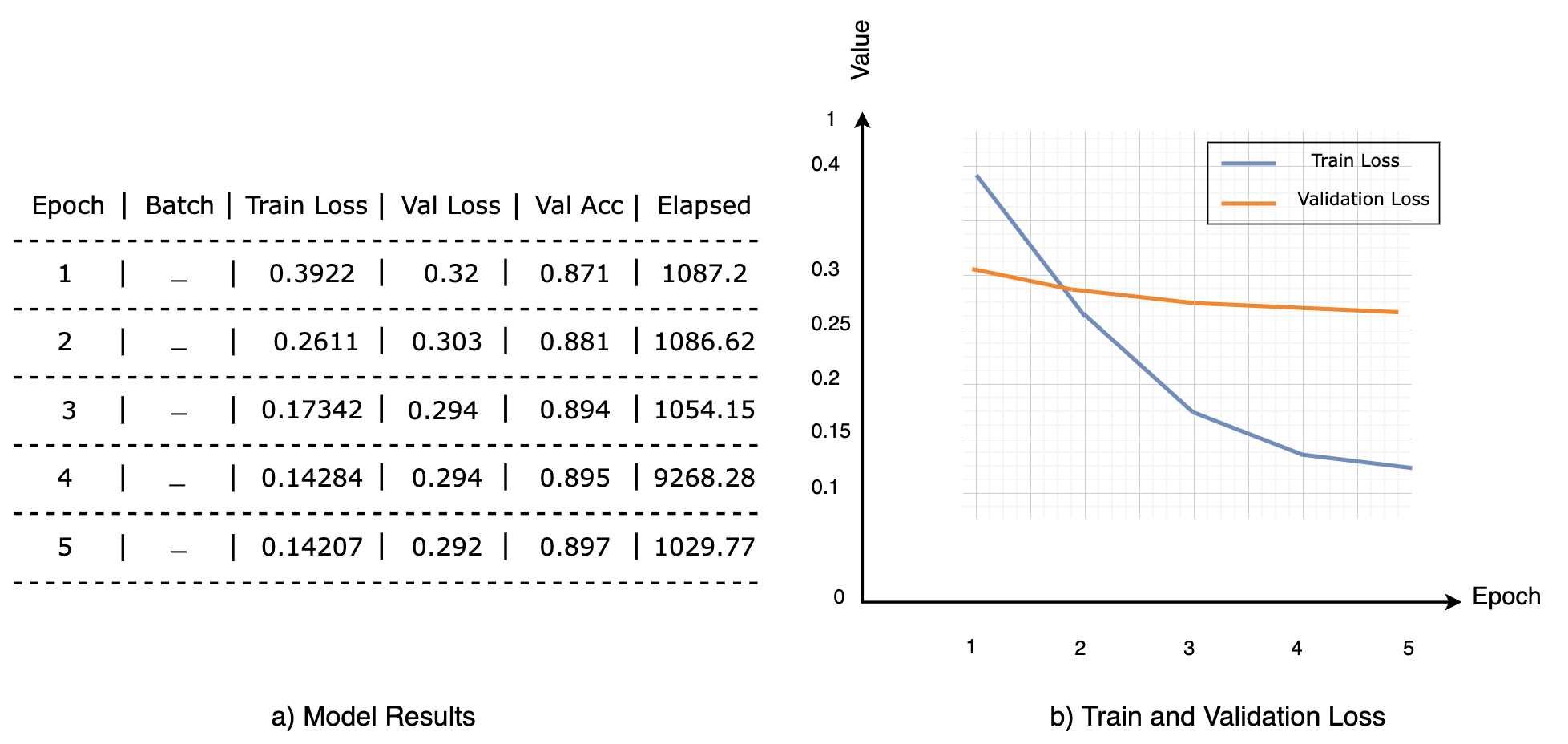}
    \caption{Comparing the loss value on the train and validation set for positive, negative and neutral sentiments based on the positive and negative feedback.}
    \label{fig:Bert05}
\end{figure}      

\bigskip

\item\textbf{Model Evaluation}
\bigskip

Generally, in machine learning and deep learning, estimating the model's performance is a critical task. This thesis discusses standard metrics used to measure the evaluation for multi-label classification and ensure the system's effective implementation. Therefore, we utilised \eqref{eq:acc} model accuracy,\eqref{eq:pres}Precision,\eqref{eq:f1score} F1-score,\eqref{eq:recall} Recall, and Area Under The Curve-Receiver Operating Characteristics (AUC - ROC) 
The following formulas represent the mathematical expressions for these performance metrics.
\bigskip

\[ Accuracy = \frac{(TP+TN)}{ (TP+TN+FP+FN)}  \label{eq:acc} \tag{1}\]
\[ Precision = \frac{TP}{ (TP+FP)}  \label{eq:pres} \tag{2}\]
\[ F1-Score = \frac{TP}{TP +\frac{1}{2}(FP+FN)} \label{eq:f1score} \tag{3}\]
\[ Recall = \frac{TP}{(TP +FN)} \label{eq:recall} \tag{4}\]

TP,  FP, FN, and TN in the mentioned formulas stand for True-Positive,  False-Positive, False-Negatives, and True-Negative, respectively. 
The model accuracy is the number of successfully labelled utterances by the tuned BERT divided by the total number of predictions. The recall is the percentage of intent labelled to relevant intents, and the F1-score is the harmonic mean of Precision and Recall.

We ran the model through a test dataset, which was not part of the training data, as we explained in section 5.4.3, based on our tuned BERT model and selected hyperparameters with an accuracy of 92.23\%.
In conclusion, the outcomes show an accurate understanding of the user's questions, providing more satisfaction to users.
The overall results of the system are as below:
\begin{table}[h!]
        \centering
        \includegraphics[scale=0.28]{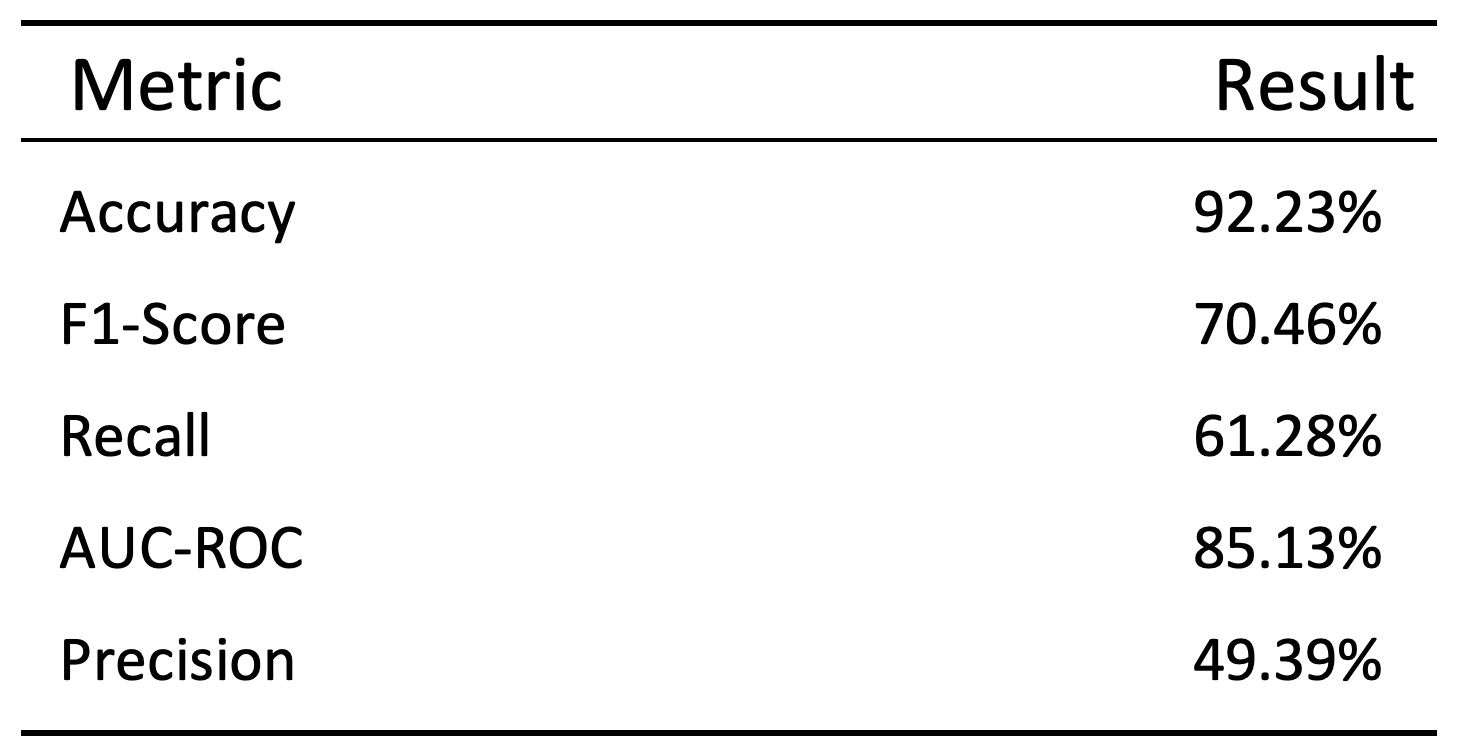}
        \caption{The model performance evaluation metrics}
        \label{tab:Performance}
\end{table}
\end{itemize}
\subsubsection{Evaluation of H2: Questionnaire and Outcomes}
This section aims to test and then confirm or reject the second hypothesis based on the obtained data throughout the experiment. H2 states that autonomous intent detection may assist domain experts and analysts by scaling services to more customers and improving customer service quality by identifying problems and challenges.
We used Google Forms to prepare the questionnaire.
All the participants are computer students or graduates working in the computer field. Therefore, they are familiar with these questions. 
    
We split participants into groups A and B. The questionnaire consists of two parts: The first part was about the participant's demographic and background, which was possessed in common between both groups. The example of the received demographic questions for groups A and B is shown in Figure 4.10. For the second section, we consider five of the same fundamental technical questions to answer about the technical issues within Microsoft windows to investigate our hypothesis. Figure 4.11 shows an example question for groups A and B.
     \begin{figure}[!ht]
    \centering
    \includegraphics[angle=0, scale=0.4]{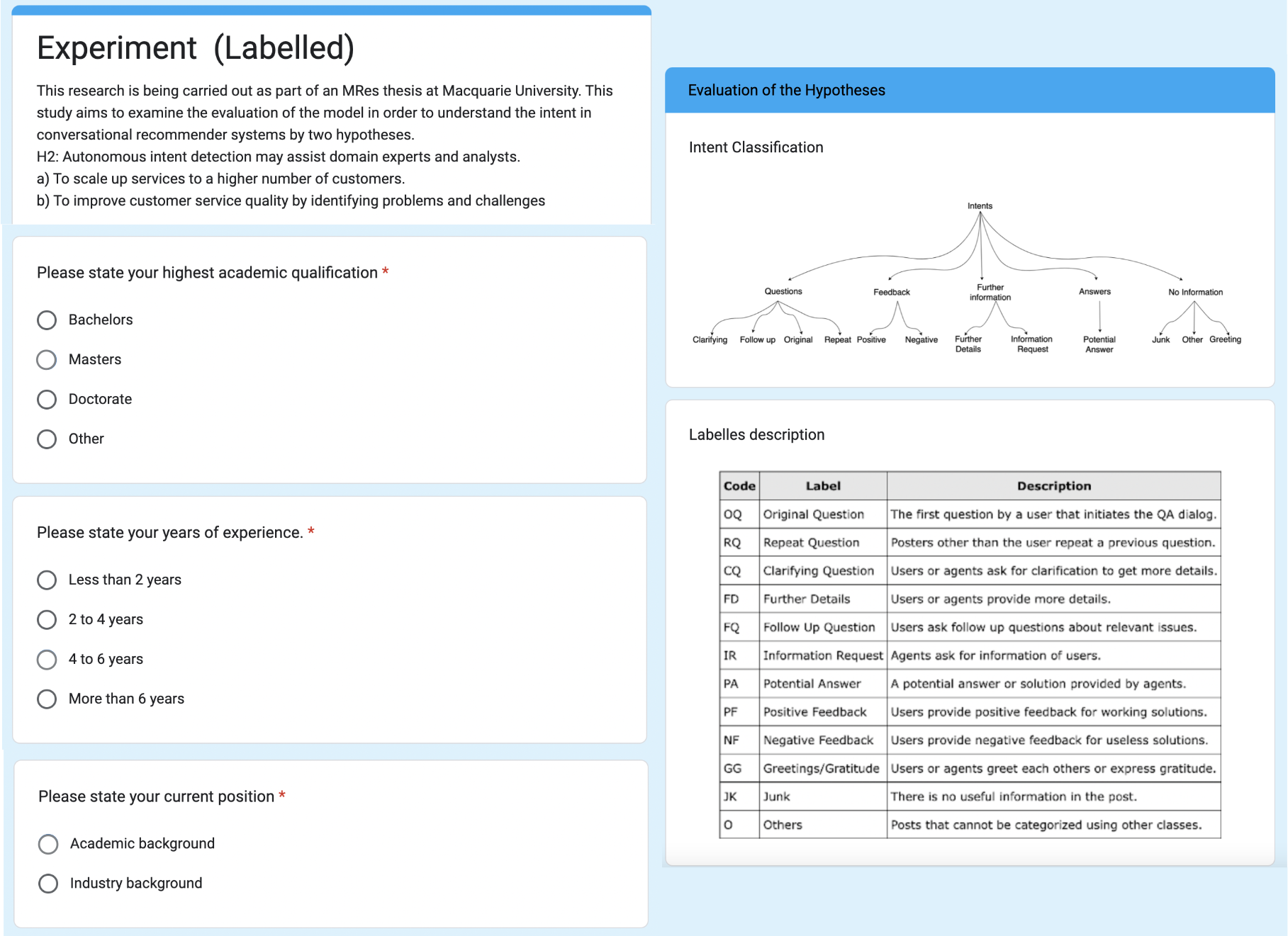}
    \caption{Demographic questions and description of the intent labels in the questionnaire.}
    \label{fig:Demographic}
\end{figure}

     \begin{figure}[!ht]
    \centering
    \includegraphics[angle=0, scale=0.45]{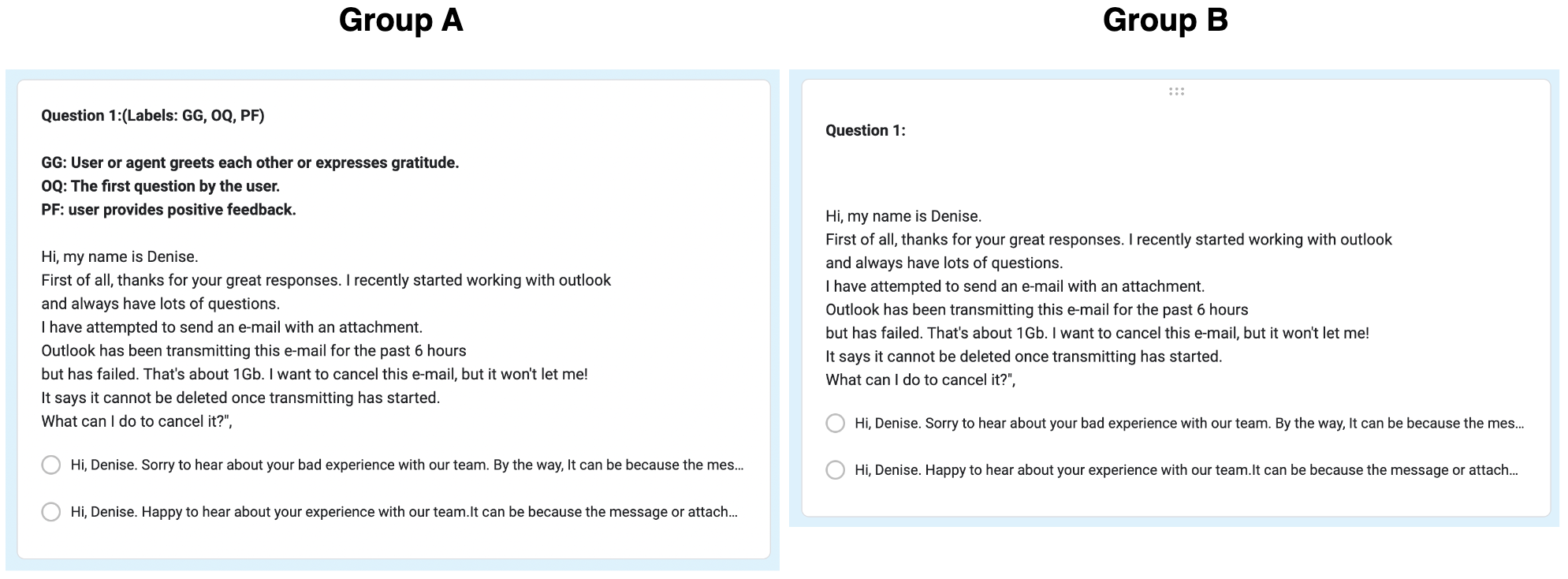}
    \caption{Example of the question for groups A and B.}
    \label{fig:question}
\end{figure}

We used the same questions from the MSDialoug conversations between users and experts and used our model to label the utterances' intents. Each question has two options for answers. From the technical point of view, and to be careful against bias and knowledge judgment of the participants, both answers are the same; the only differences are criteria based on the user intents (e.g., sentiment, feedback and greeting).
The first group received questions labelled based on our model, and from the beginning, the definition of each label was described to them. Group B received only questions without specified labels. 
Finally, to evaluate the H2,  we measure the answers' time and accuracy. Based on the acquired results. All participants in group A could finish the task in the specified time, and 50\% of the participants completed their task before the appointed time, showing the customers will receive the response faster than usual, and that proves the second part of the H1 and 66.6\% of the participants in group B could finish their tasks. Meanwhile, in group A, 86 \%, and in group B, 58.54\% of the results were correct. The last result shows that domain experts can provide more accurate responses than experts in group B. Figure 4.12 illustrate the final results.

\begin{figure}[!ht]
\centering
    \includegraphics[angle=0, scale=0.3,]{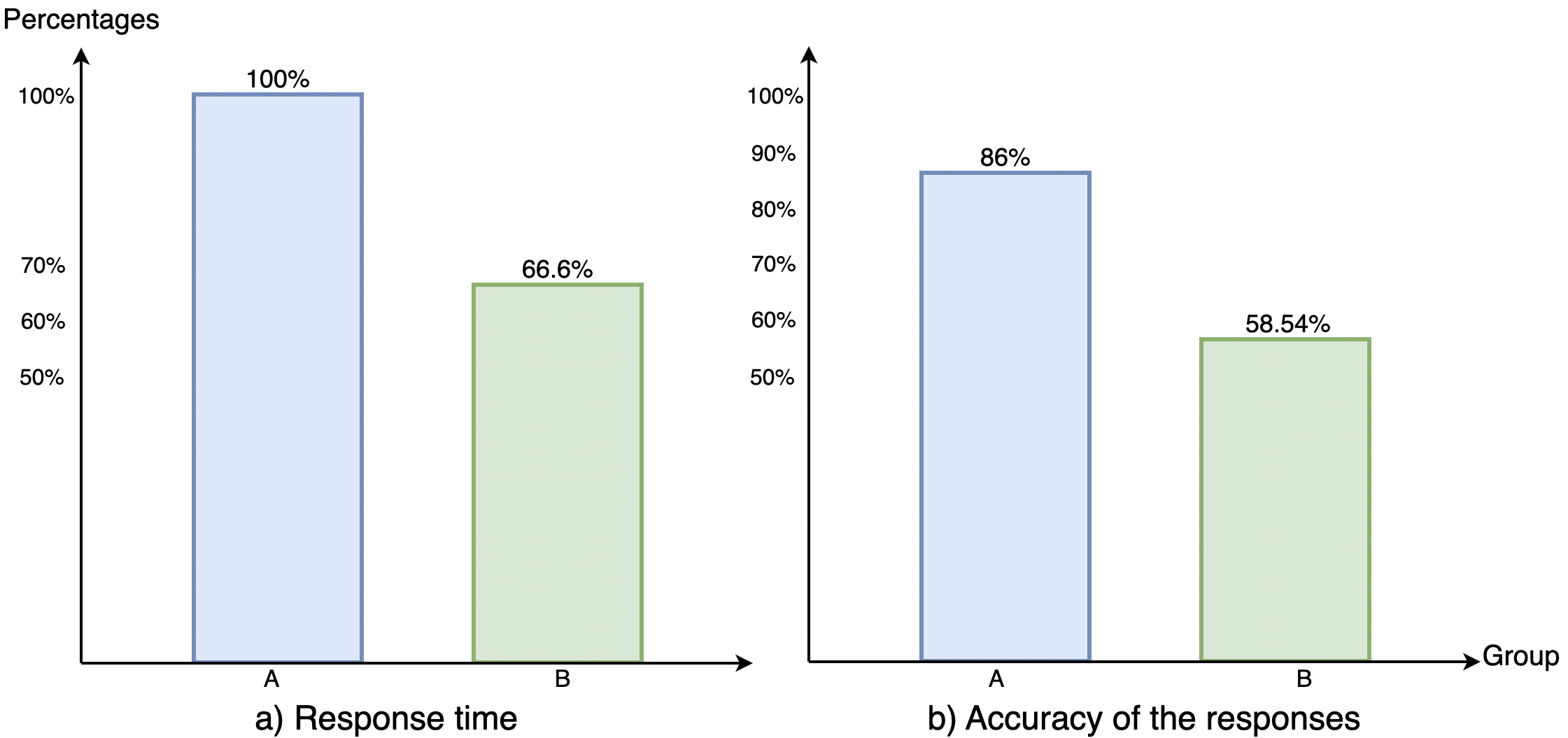}
    \caption{ Percentages of the time and the accuracy provided by participants in the specified period}
    \label{fig:Percentages}
\end{figure}

\subsubsection{Discussion of Experiment Results}

Based on the participants’ feedback, providing intent before reading about the questions assists them in having a fair idea about the user and context. The participant’s motivation and limited training about the intents and definitions significantly impact the final evaluation results. Participating in the evaluation process was entirely voluntary, which led to an acceptable engagement. The session for training did not have a fixed time, while for responding to the technical questions, we assumed 5 minutes. 
The assigned time frame for introducing the questionnaire and challenges for all the participants seems acceptable. Based on our discoveries, mainly experts in the domain provide their answers before the specified time, which shows intent labelling will help them to provide the answers in the shortest time for customers and it helps them to handle more users at the same time.

\section{Conclusion and Future Work}

\subsection{Conclusion}
Generally, understanding the preferences and intentions of users are a strategic priority for any CRS. Extracting user intent requires extensive domain knowledge and insight to use it effectively. Therefore, intent recognition is a vital component of a task-oriented conversational system.
User intent detection or recognition is a classification task based on user preferences to recommend the relevant items.

In this paper, we concentrated on understanding the user’s intent in chatbot-based conversational systems. We introduced a novel intent recognition approach to provide a qualified recommendation in the future. Hence, we developed an interactive pipeline composed of several procedures.
We introduced a pipeline to contextualise the input utterances from the human-agent conversation. The use of reverse feature engineering was, therefore, our next step. To recognise intent, we linked the contextualised input and learning model. Although performance evaluation is achieved based on different machine learning models and approaches, we used the advantages of the BERT model for this proposed approach.

\subsection{Future Research Areas}

Conversational Recommender systems enable companies to provide an intelligent platform to boost their productivity, reduce service costs, and boost customer expectations and attention. Hence, understanding the meaning behind the content is a vital part of chatbot-based conversational recommender systems. To this end, considerable research has been performed on conversational recommender systems and worked in the real world.

\subsection{leverage of Explainable Artificial Intelligence in CRSs }

Explainable artificial intelligence (XAI) ~\cite{chap5_1} is a set of techniques and approaches that help users to understand the efficiency and trustworthiness of machine learning outcomes. Broadly, they must address crucial and complex challenges in the field. Providing explanations in a form like human conversation should be a step forward in conversational recommender systems.
While numerous strategies show the technical issues in the human-machine dialogue, analysing the user perspective improves the trust in conversation. Hence, determining the explanation levels and blending the XAI in human-agent conversation will help to have a significant understanding and explanation system.

\subsection{Intent Recognition in Open-domain CRSs}

The conversational system is an operational area of investigation, started by the basic ruled-based ELIZA (mentioned in chapter 2.4) and progressed over the last few years. An open-domain conversational system is the subject of debate in the field. However, this type of chatbot does not have a good understanding of users' intent. Generally, precise analyses to identify and understand user intent are insufficient and scattered.
Hence, a complex challenge in this domain is user satisfaction prediction and extracting user intent to estimate the user's interest in interactive recommender systems.

\subsection{Extracting User's Intent-Based on Similar Session Information}

In a session-based CRS, sessions organize sequentially based on the time order, and the recommendations will generate exclusively relying on the progressed session. One of the outstanding features of a session-based conversational recommender system is performing an anonymous interaction.
By leveraging the collaborative information, try to extract the intent of users who may have matching behaviours and improve the outcomes proficiency of the recommendation in open domains.

\textbf{Acknowledgements}

- I acknowledge the Centre for Applied Artificial Intelligence at Macquarie University for funding My Master by Research project.
\bibliography{references}
\bibliographystyle{abbrv}

\end{document}